\documentclass[10pt,conference]{IEEEtran}

\usepackage{cite}
\usepackage{amsmath, amssymb, amsfonts}
\usepackage{graphicx}
\usepackage[hidelinks]{hyperref}
\usepackage{microtype}
\usepackage{url}
\usepackage{makecell}
\usepackage{multirow}
\usepackage{interval}
\intervalconfig{ soft open fences, }
\ifdefined\RefreshMintedCache
  \usepackage[cachedir=minted-cache]{minted-cache/minted}
\else
  \usepackage[cachedir=minted-cache,frozencache]{minted-cache/minted}
\fi
\usepackage[autolanguage]{numprint}
\usepackage[normalem]{ulem}
\usepackage{xspace}
\usepackage{enumitem}
\graphicspath{{figures/}}
\begin{document}

\setminted[python]{
mathescape,
linenos,
numbersep=4pt,
xleftmargin=8pt,
gobble=0,
frame=lines,
framesep=1mm,
breaklines,
fontsize=\scriptsize,
escapeinside=||
}

\setminted[cpp]{
mathescape,
linenos,
numbersep=4pt,
xleftmargin=8pt,
gobble=0,
frame=lines,
framesep=1mm,
breaklines,
fontsize=\scriptsize,
escapeinside=||
}

\newcommand{\Description}[1]{}

\newcommand{\github}{GitHub\xspace}
\newcommand{\cuda}{CUDA\xspace}
\newcommand{\triton}{Triton\xspace}
\newcommand{\cutlass}{CUTLASS\xspace}
\newcommand{\flashinfer}{FlashInfer\xspace}
\newcommand{\flashattention}{FlashAttention\xspace}
\newcommand{\deepgemm}{DeepGEMM\xspace}
\newcommand{\xformers}{xFormers\xspace}
\newcommand{\bnb}{bitsandbytes\xspace}

\newcommand{\FlashInferRawIssueCount}{1,051\xspace}
\newcommand{\FlashInferCandidateIssueCount}{390\xspace}
\newcommand{\FlashInferFinalIssueCount}{37\xspace}
\newcommand{\FlashAttentionRawIssueCount}{1,786\xspace}
\newcommand{\FlashAttentionCandidateIssueCount}{302\xspace}
\newcommand{\FlashAttentionFinalIssueCount}{28\xspace}
\newcommand{\TritonRawIssueCount}{2,205\xspace}
\newcommand{\TritonCandidateIssueCount}{815\xspace}
\newcommand{\TritonFinalIssueCount}{77\xspace}
\newcommand{\CutlassRawIssueCount}{1,931\xspace}
\newcommand{\CutlassCandidateIssueCount}{597\xspace}
\newcommand{\CutlassFinalIssueCount}{56\xspace}
\newcommand{\DeepGemmRawIssueCount}{180\xspace}
\newcommand{\DeepGemmCandidateIssueCount}{38\xspace}
\newcommand{\DeepGemmFinalIssueCount}{4\xspace}
\newcommand{\XformersRawIssueCount}{722\xspace}
\newcommand{\XformersCandidateIssueCount}{206\xspace}
\newcommand{\XformersFinalIssueCount}{19\xspace}
\newcommand{\BitsAndBytesRawIssueCount}{1,224\xspace}
\newcommand{\BitsAndBytesCandidateIssueCount}{833\xspace}
\newcommand{\BitsAndBytesFinalIssueCount}{79\xspace}
\newcommand{\RawIssueCount}{9,099\xspace}
\newcommand{\CandidateIssueCount}{3,181\xspace}
\newcommand{\FinalIssueCount}{300\xspace}
\newcommand{\finding}[3]{%
  \par\smallskip
  \noindent\hfill\fbox{\begin{minipage}{0.92\linewidth}%
    \textbf{Finding #1:} #2\par\smallskip
    \textbf{Implication:} #3
  \end{minipage}}\hfill\mbox{}%
  \par\smallskip
}

\newcommand{\SymBuildDeployment}{Build Failure\xspace}
\newcommand{\SymExecutionFailure}{Crash\xspace}
\newcommand{\SymFunctionalIncorrectness}{Incorrect Functionality\xspace}
\newcommand{\SymPerformanceDegradation}{Performance Degradation\xspace}
\newcommand{\SymProgramHang}{Hang\xspace}
\newcommand{\SymDeploymentFailure}{Deployment Failure\xspace}
\newcommand{\SymPlatformIncompatibility}{Platform Incompatibility\xspace}

\newcommand{\RcRuntimeImplementation}{Kernel Library\xspace}
\newcommand{\RcCompilerInfrastructure}{Compiler and Build System\xspace}
\newcommand{\RcComponentIntegration}{Cross-Component Integration\xspace}
\newcommand{\RcExecutionEnvironment}{Runtime Environment\xspace}
\newcommand{\RcUnresolvedLocus}{Unresolved\xspace}
\newcommand{\RcComputationalLogic}{Code Logic Error\xspace}
\newcommand{\RcNumericalSemantics}{Incorrect Numerical Computation\xspace}
\newcommand{\RcIndexing}{Incorrect Indexing\xspace}
\newcommand{\RcDataLayout}{Tensor Layout Mismatch\xspace}
\newcommand{\RcMemoryAccess}{Improper Memory Access\xspace}
\newcommand{\RcRuntimeState}{State Mismanagement\xspace}
\newcommand{\RcConcurrencyCoordination}{Concurrency Error\xspace}
\newcommand{\RcRuntimeDispatch}{Improper Kernel Dispatch\xspace}
\newcommand{\RcResourcePlanning}{Resource Mismanagement\xspace}
\newcommand{\RcInefficientExecution}{Performance Bug\xspace}
\newcommand{\RcCompilerTransformation}{Incorrect Compiler Transformation\xspace}
\newcommand{\RcBuildConfiguration}{Build Misconfiguration\xspace}
\newcommand{\RcInterfaceMismatch}{API Misuse\xspace}
\newcommand{\RcIntercomponentProtocol}{Intercomponent Protocol Fault\xspace}
\newcommand{\RcHardwarePlatform}{Hardware Incompatibility\xspace}
\newcommand{\RcSoftwareEnvironment}{Software Incompatibility\xspace}
\newcommand{\RcUnresolvedCause}{Unresolved Cause\xspace}
\newcommand{\RcBuildInfrastructure}{\RcCompilerInfrastructure}

\newcommand{\SymptomCategoryCount}{6\xspace}
\newcommand{\RootCauseCategoryCount}{15\xspace}
\newcommand{\RootCauseDimensionCount}{3\xspace}
\newcommand{\NonKernelLibraryCauseCount}{181\xspace}
\newcommand{\NonKernelLibraryCauseRatio}{60.33\%\xspace}
\newcommand{\ExecutionOutcomeIssueCount}{122\xspace}
\newcommand{\ExecutionOutcomeIssueRatio}{40.67\%\xspace}

\newcommand{\SymBuildDeploymentCount}{73\xspace}
\newcommand{\SymBuildDeploymentRatio}{24.33\%\xspace}
\newcommand{\SymBuildDeploymentFlashAttentionCount}{5\xspace}
\newcommand{\SymBuildDeploymentFlashAttentionRatio}{1.67\%\xspace}
\newcommand{\SymBuildDeploymentCutlassCount}{22\xspace}
\newcommand{\SymBuildDeploymentCutlassRatio}{7.33\%\xspace}
\newcommand{\SymBuildDeploymentBitsAndBytesCount}{5\xspace}
\newcommand{\SymBuildDeploymentBitsAndBytesRatio}{1.67\%\xspace}
\newcommand{\SymBuildDeploymentDeepGemmCount}{2\xspace}
\newcommand{\SymBuildDeploymentDeepGemmRatio}{0.67\%\xspace}
\newcommand{\SymBuildDeploymentXformersCount}{3\xspace}
\newcommand{\SymBuildDeploymentXformersRatio}{1.00\%\xspace}
\newcommand{\SymBuildDeploymentFlashInferCount}{7\xspace}
\newcommand{\SymBuildDeploymentFlashInferRatio}{2.33\%\xspace}
\newcommand{\SymBuildDeploymentTritonCount}{29\xspace}
\newcommand{\SymBuildDeploymentTritonRatio}{9.67\%\xspace}

\newcommand{\SymExecutionFailureCount}{63\xspace}
\newcommand{\SymExecutionFailureRatio}{21.00\%\xspace}
\newcommand{\SymExecutionFailureFlashAttentionCount}{6\xspace}
\newcommand{\SymExecutionFailureFlashAttentionRatio}{2.00\%\xspace}
\newcommand{\SymExecutionFailureCutlassCount}{8\xspace}
\newcommand{\SymExecutionFailureCutlassRatio}{2.67\%\xspace}
\newcommand{\SymExecutionFailureBitsAndBytesCount}{16\xspace}
\newcommand{\SymExecutionFailureBitsAndBytesRatio}{5.33\%\xspace}
\newcommand{\SymExecutionFailureDeepGemmCount}{1\xspace}
\newcommand{\SymExecutionFailureDeepGemmRatio}{0.33\%\xspace}
\newcommand{\SymExecutionFailureXformersCount}{4\xspace}
\newcommand{\SymExecutionFailureXformersRatio}{1.33\%\xspace}
\newcommand{\SymExecutionFailureFlashInferCount}{18\xspace}
\newcommand{\SymExecutionFailureFlashInferRatio}{6.00\%\xspace}
\newcommand{\SymExecutionFailureTritonCount}{10\xspace}
\newcommand{\SymExecutionFailureTritonRatio}{3.33\%\xspace}

\newcommand{\SymFunctionalIncorrectnessCount}{53\xspace}
\newcommand{\SymFunctionalIncorrectnessRatio}{17.67\%\xspace}
\newcommand{\SymFunctionalIncorrectnessFlashAttentionCount}{7\xspace}
\newcommand{\SymFunctionalIncorrectnessFlashAttentionRatio}{2.33\%\xspace}
\newcommand{\SymFunctionalIncorrectnessCutlassCount}{13\xspace}
\newcommand{\SymFunctionalIncorrectnessCutlassRatio}{4.33\%\xspace}
\newcommand{\SymFunctionalIncorrectnessBitsAndBytesCount}{9\xspace}
\newcommand{\SymFunctionalIncorrectnessBitsAndBytesRatio}{3.00\%\xspace}
\newcommand{\SymFunctionalIncorrectnessDeepGemmCount}{1\xspace}
\newcommand{\SymFunctionalIncorrectnessDeepGemmRatio}{0.33\%\xspace}
\newcommand{\SymFunctionalIncorrectnessXformersCount}{1\xspace}
\newcommand{\SymFunctionalIncorrectnessXformersRatio}{0.33\%\xspace}
\newcommand{\SymFunctionalIncorrectnessFlashInferCount}{6\xspace}
\newcommand{\SymFunctionalIncorrectnessFlashInferRatio}{2.00\%\xspace}
\newcommand{\SymFunctionalIncorrectnessTritonCount}{16\xspace}
\newcommand{\SymFunctionalIncorrectnessTritonRatio}{5.33\%\xspace}

\newcommand{\SymPerformanceDegradationCount}{23\xspace}
\newcommand{\SymPerformanceDegradationRatio}{7.67\%\xspace}
\newcommand{\SymPerformanceDegradationFlashAttentionCount}{4\xspace}
\newcommand{\SymPerformanceDegradationFlashAttentionRatio}{1.33\%\xspace}
\newcommand{\SymPerformanceDegradationCutlassCount}{5\xspace}
\newcommand{\SymPerformanceDegradationCutlassRatio}{1.67\%\xspace}
\newcommand{\SymPerformanceDegradationBitsAndBytesCount}{2\xspace}
\newcommand{\SymPerformanceDegradationBitsAndBytesRatio}{0.67\%\xspace}
\newcommand{\SymPerformanceDegradationDeepGemmCount}{0\xspace}
\newcommand{\SymPerformanceDegradationDeepGemmRatio}{0.00\%\xspace}
\newcommand{\SymPerformanceDegradationXformersCount}{3\xspace}
\newcommand{\SymPerformanceDegradationXformersRatio}{1.00\%\xspace}
\newcommand{\SymPerformanceDegradationFlashInferCount}{2\xspace}
\newcommand{\SymPerformanceDegradationFlashInferRatio}{0.67\%\xspace}
\newcommand{\SymPerformanceDegradationTritonCount}{7\xspace}
\newcommand{\SymPerformanceDegradationTritonRatio}{2.33\%\xspace}

\newcommand{\SymProgramHangCount}{6\xspace}
\newcommand{\SymProgramHangRatio}{2.00\%\xspace}
\newcommand{\SymProgramHangFlashAttentionCount}{1\xspace}
\newcommand{\SymProgramHangFlashAttentionRatio}{0.33\%\xspace}
\newcommand{\SymProgramHangCutlassCount}{2\xspace}
\newcommand{\SymProgramHangCutlassRatio}{0.67\%\xspace}
\newcommand{\SymProgramHangBitsAndBytesCount}{0\xspace}
\newcommand{\SymProgramHangBitsAndBytesRatio}{0.00\%\xspace}
\newcommand{\SymProgramHangDeepGemmCount}{0\xspace}
\newcommand{\SymProgramHangDeepGemmRatio}{0.00\%\xspace}
\newcommand{\SymProgramHangXformersCount}{0\xspace}
\newcommand{\SymProgramHangXformersRatio}{0.00\%\xspace}
\newcommand{\SymProgramHangFlashInferCount}{1\xspace}
\newcommand{\SymProgramHangFlashInferRatio}{0.33\%\xspace}
\newcommand{\SymProgramHangTritonCount}{2\xspace}
\newcommand{\SymProgramHangTritonRatio}{0.67\%\xspace}

\newcommand{\SymDeploymentFailureCount}{82\xspace}
\newcommand{\SymDeploymentFailureRatio}{27.33\%\xspace}
\newcommand{\SymDeploymentFailureFlashAttentionCount}{5\xspace}
\newcommand{\SymDeploymentFailureFlashAttentionRatio}{1.67\%\xspace}
\newcommand{\SymDeploymentFailureCutlassCount}{6\xspace}
\newcommand{\SymDeploymentFailureCutlassRatio}{2.00\%\xspace}
\newcommand{\SymDeploymentFailureBitsAndBytesCount}{47\xspace}
\newcommand{\SymDeploymentFailureBitsAndBytesRatio}{15.67\%\xspace}
\newcommand{\SymDeploymentFailureDeepGemmCount}{0\xspace}
\newcommand{\SymDeploymentFailureDeepGemmRatio}{0.00\%\xspace}
\newcommand{\SymDeploymentFailureXformersCount}{8\xspace}
\newcommand{\SymDeploymentFailureXformersRatio}{2.67\%\xspace}
\newcommand{\SymDeploymentFailureFlashInferCount}{3\xspace}
\newcommand{\SymDeploymentFailureFlashInferRatio}{1.00\%\xspace}
\newcommand{\SymDeploymentFailureTritonCount}{13\xspace}
\newcommand{\SymDeploymentFailureTritonRatio}{4.33\%\xspace}

\newcommand{\RcRuntimeImplementationCount}{119\xspace}
\newcommand{\RcRuntimeImplementationRatio}{39.67\%\xspace}

\newcommand{\RcCompilerInfrastructureCount}{81\xspace}
\newcommand{\RcCompilerInfrastructureRatio}{27.00\%\xspace}

\newcommand{\RcExecutionEnvironmentCount}{100\xspace}
\newcommand{\RcExecutionEnvironmentRatio}{33.33\%\xspace}

\newcommand{\RcComputationalLogicCount}{22\xspace}
\newcommand{\RcComputationalLogicRatio}{7.33\%\xspace}

\newcommand{\RcNumericalSemanticsCount}{8\xspace}
\newcommand{\RcNumericalSemanticsRatio}{2.67\%\xspace}

\newcommand{\RcIndexingCount}{4\xspace}
\newcommand{\RcIndexingRatio}{1.33\%\xspace}

\newcommand{\RcDataLayoutCount}{12\xspace}
\newcommand{\RcDataLayoutRatio}{4.00\%\xspace}

\newcommand{\RcMemoryAccessCount}{7\xspace}
\newcommand{\RcMemoryAccessRatio}{2.33\%\xspace}

\newcommand{\RcRuntimeStateCount}{2\xspace}
\newcommand{\RcRuntimeStateRatio}{0.67\%\xspace}

\newcommand{\RcConcurrencyCoordinationCount}{5\xspace}
\newcommand{\RcConcurrencyCoordinationRatio}{1.67\%\xspace}

\newcommand{\RcRuntimeDispatchCount}{10\xspace}
\newcommand{\RcRuntimeDispatchRatio}{3.33\%\xspace}

\newcommand{\RcResourcePlanningCount}{6\xspace}
\newcommand{\RcResourcePlanningRatio}{2.00\%\xspace}

\newcommand{\RcInefficientExecutionCount}{17\xspace}
\newcommand{\RcInefficientExecutionRatio}{5.67\%\xspace}

\newcommand{\RcCompilerTransformationCount}{46\xspace}
\newcommand{\RcCompilerTransformationRatio}{15.33\%\xspace}

\newcommand{\RcBuildConfigurationCount}{35\xspace}
\newcommand{\RcBuildConfigurationRatio}{11.67\%\xspace}

\newcommand{\RcInterfaceMismatchCount}{26\xspace}
\newcommand{\RcInterfaceMismatchRatio}{8.67\%\xspace}

\newcommand{\RcHardwarePlatformCount}{24\xspace}
\newcommand{\RcHardwarePlatformRatio}{8.00\%\xspace}

\newcommand{\RcSoftwareEnvironmentCount}{76\xspace}
\newcommand{\RcSoftwareEnvironmentRatio}{25.33\%\xspace}

\newcommand{\ReviewSampleSize}{300\xspace}
\newcommand{\IndependentSymptomKappa}{0.969\xspace}
\newcommand{\IndependentRootCauseKappa}{0.881\xspace}
\newcommand{\IndependentJointKappa}{0.87\xspace}
\newcommand{\FinalSymptomMatches}{300/300\xspace}
\newcommand{\FinalRootCauseMatches}{300/300\xspace}
\newcommand{\FinalRootDimensionMatches}{300/300\xspace}
\newcommand{\FinalSymptomAgreement}{100\%\xspace}
\newcommand{\FinalRootCauseAgreement}{100\%\xspace}
\newcommand{\FinalRootDimensionAgreement}{100\%\xspace}
\newcommand{\FinalSymptomRootJointMatches}{300/300\xspace}
\newcommand{\FinalSymptomRootJointAgreement}{100\%\xspace}
\newcommand{\FinalCertaintyMatches}{240/300\xspace}
\newcommand{\FinalCertaintyAgreement}{80.0\%\xspace}
\newcommand{\FinalFixMatches}{172/300\xspace}
\newcommand{\FinalFixAgreement}{57.3\%\xspace}
\newcommand{\FinalOtherFieldDisagreementCases}{157\xspace}

\title{An Empirical Study of Issues in GPU Kernel Libraries for LLM Systems}

\DeclareRobustCommand*{\IEEEauthorrefmark}[1]{\raisebox{0pt}[0pt][0pt]{\textsuperscript{\footnotesize\ensuremath{#1}}}}

\author{
    \IEEEauthorblockN{
        Yanjie Gao\IEEEauthorrefmark{1},
        Jiayi Bi\IEEEauthorrefmark{2},
        Haoxiang Lin\IEEEauthorrefmark{1},
        Ling Feng\IEEEauthorrefmark{2}
    }
    \IEEEauthorblockA{
        \IEEEauthorrefmark{1}Microsoft Research Asia, Beijing, China \qquad
        \IEEEauthorrefmark{2}Tsinghua University, Beijing, China
    }
    \IEEEauthorblockA{
        Email: \{yanjga,haoxlin\}@microsoft.com, \{bijy22,fengling\}@mails.tsinghua.edu.cn
    }
}

\maketitle

\begin{abstract}

Behind the expanding use of large language models (LLMs) across application domains lies a growing dependence on specialized GPU kernel libraries.
These libraries underpin the correctness and efficiency of training and inference across diverse GPU platforms, implementing attention, GEMM, quantization, and low-precision operators while managing tensor shapes, memory layouts, code generation, and hardware-specific dispatch.
Problems at this layer can silently corrupt outputs, hang GPUs, prevent compilation, introduce deployment incompatibilities, or degrade performance,
thereby reducing developer productivity and wasting expensive GPU resources.

We present the first comprehensive empirical study of issues in GPU kernel libraries for LLM systems.
We manually analyzed \FinalIssueCount high-quality GitHub issue reports, their discussions, and available fixes from seven representative libraries: \flashinfer, \flashattention, \triton, \cutlass, \deepgemm, \xformers, and \bnb.
The analysis characterizes symptoms, root causes, fixes, and testing and debugging practices.
Our main findings are:
(1) \NonKernelLibraryCauseRatio{} of the sampled issues originate in the \RcCompilerInfrastructure{} or \RcExecutionEnvironment{} dimensions, so validation must cover the deployment and execution stack, not just kernel code.
(2) Kernel specialization imposes implicit correctness constraints on tensor shapes, metadata layouts, memory aliasing, and hardware resources.
Violations can produce incorrect outputs or launch failures even for supported operations, highlighting the need for explicit contract validation and boundary-case testing.
(3) Current testing and debugging practices are often insufficient for identifying low-precision numerical problems and hardware-specific execution failures.
We outline research opportunities and tooling directions to support future LLM GPU kernel development.

\end{abstract}


\section{Introduction}

In recent years, large language models (LLMs) have become a core component of modern software systems.
Their applications range from chatbots and programming assistants to search, education, and enterprise software.
Training these models and serving their requests require large-scale GPU infrastructure, where tensor operations account for much of the runtime cost.

Developers rely on specialized GPU kernel libraries, such as \flashattention, \triton, and \cutlass, to execute these workloads efficiently~\cite{Dao2022FlashAttention,FlashInferRepo,Tillet2019Triton,CUTLASSRepo,DeepGEMMRepo,XFormersRepo,BitsAndBytesRepo}.
These libraries provide optimized implementations of attention, GEMM, and quantization.
Their implementations incorporate low-precision arithmetic and hardware-specific code generation to support diverse tensor shapes and layouts, data types, and GPU architectures.

Failures in these libraries can cause crashes, hangs, incorrect outputs, deployment failures, and performance regressions, reducing developer productivity and wasting expensive GPU resources. 
For example, a GPU kernel that produces incorrect results may silently corrupt generated tokens without raising a runtime exception;
a synchronization bug may hang a long-running inference job, while a packaging or CUDA incompatibility may prevent deployment before a kernel is even tested. 
Worse still, these failures' impacts can extend further than those of failures in ordinary application code because a single kernel path may be reused across many models, serving requests, GPUs, and downstream systems.

The cause of such a failure is often difficult to isolate.
Its visible symptom may depend on a particular combination of tensor shape/layout, data type, GPU architecture, CUDA or driver version, generated code path, and downstream framework.
These challenges underscore the need for a systematic understanding of GPU kernel failures and their underlying causes, which can inform efforts to prevent, detect, debug, and resolve such failures.
As LLMs become integral to modern software, the reliability of the GPU kernel libraries supporting their training and inference deserves a high degree of attention from software engineering researchers and practitioners.

Prior empirical research has examined problems in machine-learning and deep-learning software from different perspectives.
Shen et al.~\cite{Shen2021} characterized defects in deep-learning compilers, while Humbatova et al.~\cite{Humbatova2020DLFaults} developed a fault taxonomy for deep-learning applications.
Gao et al.~\cite{Gao2023} investigated quality issues in a deep-learning platform.
More recently, a study of LLM training systems highlighted failures involving parallel execution, configuration, and external dependencies~\cite{Gao2025}.
These studies explain many problems in applications, frameworks, compilers, and end-to-end training systems, but the GPU kernel library layer remains largely unexplored.
Diagnosis at this layer requires reasoning about tensor shapes and layouts, vectorized memory access, synchronization, low-precision semantics, code generation, and architecture-specific kernel dispatch.

This paper presents the first comprehensive empirical study of issues in GPU kernel libraries for LLM systems.
We manually analyzed 300 high-quality GitHub issue reports from seven representative libraries: \flashinfer~\cite{FlashInferRepo}, \flashattention~\cite{Dao2022FlashAttention}, \triton~\cite{Tillet2019Triton}, \cutlass~\cite{CUTLASSRepo}, \deepgemm~\cite{DeepGEMMRepo}, \xformers~\cite{XFormersRepo}, and \bnb~\cite{BitsAndBytesRepo}.
Our analysis draws on issue titles, descriptions, discussions, diagnostic evidence, and available fixes.
We classified the reports into \SymptomCategoryCount symptom categories and identified \RootCauseCategoryCount root-cause categories, which were further organized into \RootCauseDimensionCount dimensions.
We also examined typical fixes and the testing and debugging practices documented in the reports.
The goal is to provide a systematic account of failures in GPU kernel libraries that can inform the development of more reliable and efficient LLM systems.

Specifically, this study addresses three research questions:
\begin{enumerate}[
  label=\textbf{RQ\arabic*:},
  leftmargin=!,
  labelwidth=2.5em,
  labelsep=0.5em,
  align=left
]
  \item
  \textbf{What are the common symptoms of issues in GPU kernel libraries for LLM systems?}
  We examine the titles and initial descriptions of the 300 issue reports to identify their observable symptoms.
  The issues fall into \SymptomCategoryCount categories: \SymDeploymentFailure (\SymDeploymentFailureRatio), \SymBuildDeployment (\SymBuildDeploymentRatio), \SymExecutionFailure (\SymExecutionFailureRatio), \SymFunctionalIncorrectness (\SymFunctionalIncorrectnessRatio), \SymPerformanceDegradation (\SymPerformanceDegradationRatio), and \SymProgramHang (\SymProgramHangRatio).
  Section~\ref{sec:symptoms} presents the symptom analysis.

  \item
  \textbf{What causes these issues, and how are they typically fixed?}
  Issue discussions, code fixes, and maintainer conclusions provide the evidence for identifying root causes and summarizing typical fixes.
  We identify \RootCauseCategoryCount root-cause categories across \RootCauseDimensionCount dimensions: \RcRuntimeImplementation (\RcRuntimeImplementationRatio), \RcExecutionEnvironment{} (\RcExecutionEnvironmentRatio), and \RcCompilerInfrastructure (\RcCompilerInfrastructureRatio).
  These causes often reflect rapid changes in hardware and software, complex compiler and package dependencies, and specialization-specific assumptions about tensor shapes and layouts, data types, and execution resources.
  The root causes and corresponding fixes are detailed in Section~\ref{sec:rootcause}.

  \item
  \textbf{What testing and debugging practices are commonly used?}
  We review issue discussions, reproduction scripts, diagnostic output, and relevant reference materials to examine how developers test and debug GPU kernels.
  The analysis reveals that current practices are often insufficient for diagnosing low-precision numerical errors and hardware-specific execution failures.
  Section~\ref{sec:evidence} summarizes these practices and their limitations.

\end{enumerate}

The findings point to research opportunities and tooling improvements for GPU kernel libraries used by LLM systems.
We discuss their implications alongside the corresponding findings in Section~\ref{sec:discussion}.

In summary, this paper makes the following contributions:
\begin{enumerate}
  \item We present the first comprehensive empirical study of issues in GPU kernel libraries for LLM systems.
  Based on 300 high-quality issue reports from seven representative libraries, the study characterizes symptoms, root causes, typical fixes, and testing and debugging practices.

  \item We draw practical implications from the findings and suggest improvements to GPU kernel libraries.
  These suggestions aim to improve developer productivity and the reliability and efficiency of LLM systems.
\end{enumerate}

\section{Background}\label{sec:background}

Large language model (LLM) applications rely on training and inference systems that execute large tensor programs on GPUs.
Python is commonly used for model training, evaluation, and serving because it offers flexible APIs and a broad ecosystem.
The performance-critical operators, however, are handled by specialized libraries that supply handwritten GPU kernels, compiler-generated kernels, and reusable CUDA templates.
Examples include \flashattention~\cite{Dao2022FlashAttention}, \flashinfer~\cite{FlashInferRepo}, \triton~\cite{Tillet2019Triton}, \cutlass~\cite{CUTLASSRepo}, \deepgemm~\cite{DeepGEMMRepo}, \xformers~\cite{XFormersRepo}, and \bnb~\cite{BitsAndBytesRepo}.
This division allows LLM systems to offer high-level abstractions while specialized GPU kernels meet their latency, throughput, memory capacity, and numerical-precision requirements on modern GPUs.

GPU kernel libraries sit between model operators and the hardware that executes them.
Attention libraries optimize prefill, decode, sliding-window attention, paged KV-cache layouts, causal masking, and variable sequence lengths.
GEMM libraries and domain-specific languages (DSLs) provide access to tensor-core instructions, tiling strategies, shared-memory layouts, TMA transfers, warpgroup MMA, and low-precision formats.
Quantization libraries support weight-only and activation-aware schemes in formats such as INT8, INT4, FP8, and FP4.

A framework or serving engine may invoke these libraries indirectly during execution.
Which kernel actually runs depends on the tensor shape, layout, and data type, as well as the CUDA version and GPU architecture.
The selected path may use handwritten CUDA code, generated Triton code, CuTe templates, or packaged binaries.
Users may work entirely through high-level Python APIs, yet their programs remain subject to the contracts and failure modes of the underlying kernel libraries.

Before an operator can run, its kernel library must be installed or built for the framework and environment.
Figure~\ref{fig:kernel-code} illustrates the subsequent library-internal workflow with a simplified attention example.
The library first validates the input tensors (line 3).
Next, it determines the GPU architecture and constructs a dispatch configuration from the data type, head dimension, causal mode, and target architecture (lines 5--7).
If no prebuilt version of the selected implementation is available, the library compiles a specialization (lines 9--10).
The selected kernel is then launched, and its result is returned (line 12).

\begin{figure}[!t]
\begin{minted}{python}
def attention(q, k, v, causal):
    # 1. Check the operator contract.
    check_shape_stride_dtype(q, k, v)
    # 2. Select a kernel for the input and platform.
    target = get_gpu_architecture()
    config = (q.dtype, q.shape[-1], causal, target)
    kernel = dispatch_table.lookup(config)
    # 3. Build a specialization when needed.
    if not kernel.is_prebuilt:
        kernel = jit_compile(kernel, config)
    # 4. Launch and return the device result.
    return kernel.launch(q, k, v)
\end{minted}
\caption{Simplified workflow for an attention operator within a GPU kernel library: input validation, kernel selection, compilation, and execution.}
\label{fig:kernel-code}
\vspace{-8pt}
\end{figure}

An issue in a GPU kernel library can arise at any of these stages.
Package or binary incompatibilities can prevent installation.
Validation may fail to enforce an input contract, and dispatch may select an incorrect implementation or find no suitable one.
JIT compilation may fail before execution begins, while device execution may crash, produce incorrect results, or run inefficiently.
These library-level responsibilities distinguish GPU kernel libraries from complete systems such as Megatron-LM~\cite{shoeybi2019megatron}, vLLM~\cite{VLLMRepo}, and SGLang~\cite{SGLangRepo}.
Those systems additionally handle model loading, scheduling, distributed execution, and service-level logic.

\section{Methodology}\label{sec:methodology}

\subsection{Study Subjects}

LLM development and deployment increasingly rely on GPU kernel libraries, whose specialized implementations pose challenges in correctness, compatibility, and efficiency.
Problems in these libraries can propagate across downstream training and inference systems, wasting GPU resources and reducing developer productivity.
Examining reported issues is therefore critical for improving the quality and reliability of LLM training and inference.

GitHub issues from GPU kernel library repositories provide the empirical basis for our study.
To the best of our knowledge, prior empirical studies~\cite{Zhang2018,Zhang2019,Shen2021,Cao2022,Gao2023,Gao2025} have not specifically investigated issues in GPU kernel libraries for LLM systems.
Our objectives are to characterize the issues' symptoms, identify their root causes, analyze the corresponding fixes, and examine testing and debugging practices.
The study covers seven representative open-source repositories.
Among them, \flashattention, \flashinfer, and \xformers provide attention and LLM operator kernels, while \cutlass and \deepgemm provide GEMM and low-precision matrix kernels.
The remaining two repositories serve distinct roles: \triton provides GPU compilation and code generation, and \bnb provides quantization kernels and CUDA packaging.
Together, these projects span complementary parts of the LLM GPU kernel ecosystem.
Complete serving engines fall outside our scope because their issue trackers also encompass model-loading, scheduling, distributed-runtime, and service-level problems, which are not within the scope of this study. 

These libraries are extensively used in both industry and academia.
Research on efficient attention and GPU programming gave rise to \flashattention and \triton, respectively.
Meanwhile, \cutlass, \flashinfer, \deepgemm, \xformers, and \bnb offer reusable CUDA templates, inference kernels, low-precision GEMM, attention operators, and quantization kernels.
Their integration into downstream systems further illustrates their practical relevance.
For training, Megatron-LM, PyTorch~\cite{pytorch}, and Transformers~\cite{transformersgit} use or expose integrations with optimized attention, \triton/CUDA kernels, \xformers operators, and \bnb quantization.
For inference, vLLM integrates \flashattention, \flashinfer, \xformers, \triton, and \bnb backends, whereas SGLang integrates \flashinfer, \flashattention, \triton, and \deepgemm backends.

\subsection{Data Collection}

\begin{table}[!t]
\centering
\caption{Dataset by repository.}
\label{tab:subjects}
\renewcommand{\arraystretch}{1.2}
\begin{tabular}{|l|r|r|r|}
\hline
\bf Repository & \bf Raw & \bf Candidate & \bf Final \cr
\hline
FlashInfer & \FlashInferRawIssueCount & \FlashInferCandidateIssueCount & \FlashInferFinalIssueCount \cr
FlashAttention & \FlashAttentionRawIssueCount & \FlashAttentionCandidateIssueCount & \FlashAttentionFinalIssueCount \cr
Triton & \TritonRawIssueCount & \TritonCandidateIssueCount & \TritonFinalIssueCount \cr
CUTLASS & \CutlassRawIssueCount & \CutlassCandidateIssueCount & \CutlassFinalIssueCount \cr
DeepGEMM & \DeepGemmRawIssueCount & \DeepGemmCandidateIssueCount & \DeepGemmFinalIssueCount \cr
xFormers & \XformersRawIssueCount & \XformersCandidateIssueCount & \XformersFinalIssueCount \cr
bitsandbytes & \BitsAndBytesRawIssueCount & \BitsAndBytesCandidateIssueCount & \BitsAndBytesFinalIssueCount \cr
\hline
Total & \RawIssueCount & \CandidateIssueCount & \FinalIssueCount \cr
\hline
\end{tabular}
\vspace{-8pt}
\end{table}

GitHub is a widely used source for empirical investigations of software issues.
Following prior research~\cite{8115662}, we collected reports from the official repositories of the selected libraries to examine problems encountered during their development and use.
These repositories remain active, with maintainers continually refining GPU kernel implementations, compiler support, and platform compatibility.
Users contribute not only failure reports but also reproduction details, diagnostic information, and input data to the resolution process.
Compared with commits, issue reports provide a richer account of a problem, spanning its initial description, subsequent discussion, diagnostic evidence, and developer resolution~\cite{8115662}.
These contexts make GitHub issues particularly suitable for examining symptoms, identifying root causes, and analyzing fixes.

Using the GitHub REST API~\cite{githubrestapi}, we collected all open and closed issues from the seven repositories between December 26, 2015 and July 10, 2026.
For each report, we downloaded its title, description, labels, state, timestamps, comments, and linked GitHub references.
The initial collection comprised \RawIssueCount issues.
To focus on concrete and complete failures of existing functionality, we first removed open issues and retained only closed reports meeting this criterion.
Repository-specific labels, considered alongside report content, informed the filtering process.
Screening began by locating issues whose titles contained explicit failure-related terms, including ``bug'', ``error'', ``fail'', ``wrong'', ``crash'', ``hang'', ``regression'', ``performance'', and ``numerical issue''.
We further eliminated placeholders, support questions, feature or capability requests, documentation and maintenance tasks, broad tracking issues, reports lacking explicit failure evidence, and duplicates.
Labels such as ``question'', ``feature request'', ``enhancement'', ``documentation/docs'', ``roadmap'', ``tracking'', ``help wanted'', ``good first issue'', ``chore'', and ``style'' helped identify reports outside the study scope.
The screening process yielded a candidate issue pool of \CandidateIssueCount closed reports.

\begin{table*}[!t]
  \centering
  \caption{Classification of Common Symptoms.}
  \label{tab:taxonomy}
  \scriptsize
  \setlength{\tabcolsep}{1.8pt}
  \resizebox{0.95\textwidth}{!}{%
  \begin{tabular}{|l|c|c|c|c|c|c|c|c|c|c|c|c|c|c|c|c|}
    \hline
    \multirow{2}{*}{\bf Category}
    & \multicolumn{2}{c|}{\bf Total}
    & \multicolumn{2}{c|}{\bf FlashInfer} & \multicolumn{2}{c|}{\bf FlashAttention}
    & \multicolumn{2}{c|}{\bf Triton} & \multicolumn{2}{c|}{\bf CUTLASS}
    & \multicolumn{2}{c|}{\bf DeepGEMM} & \multicolumn{2}{c|}{\bf xFormers}
    & \multicolumn{2}{c|}{\bf bitsandbytes} \\
    \cline{2-17}
    & \bf No. & \bf Ratio & \bf No. & \bf Ratio & \bf No. & \bf Ratio
    & \bf No. & \bf Ratio & \bf No. & \bf Ratio & \bf No. & \bf Ratio
    & \bf No. & \bf Ratio & \bf No. & \bf Ratio \\
    \hline
    \SymDeploymentFailure{}       & \SymDeploymentFailureCount & \SymDeploymentFailureRatio & \SymDeploymentFailureFlashInferCount & \SymDeploymentFailureFlashInferRatio & \SymDeploymentFailureFlashAttentionCount & \SymDeploymentFailureFlashAttentionRatio & \SymDeploymentFailureTritonCount & \SymDeploymentFailureTritonRatio & \SymDeploymentFailureCutlassCount & \SymDeploymentFailureCutlassRatio & \SymDeploymentFailureDeepGemmCount & \SymDeploymentFailureDeepGemmRatio & \SymDeploymentFailureXformersCount & \SymDeploymentFailureXformersRatio & \SymDeploymentFailureBitsAndBytesCount & \SymDeploymentFailureBitsAndBytesRatio \\
    \hline
    \SymBuildDeployment{}         & \SymBuildDeploymentCount & \SymBuildDeploymentRatio & \SymBuildDeploymentFlashInferCount & \SymBuildDeploymentFlashInferRatio & \SymBuildDeploymentFlashAttentionCount & \SymBuildDeploymentFlashAttentionRatio & \SymBuildDeploymentTritonCount & \SymBuildDeploymentTritonRatio & \SymBuildDeploymentCutlassCount & \SymBuildDeploymentCutlassRatio & \SymBuildDeploymentDeepGemmCount & \SymBuildDeploymentDeepGemmRatio & \SymBuildDeploymentXformersCount & \SymBuildDeploymentXformersRatio & \SymBuildDeploymentBitsAndBytesCount & \SymBuildDeploymentBitsAndBytesRatio \\
    \hline
    \SymExecutionFailure{}        & \SymExecutionFailureCount & \SymExecutionFailureRatio & \SymExecutionFailureFlashInferCount & \SymExecutionFailureFlashInferRatio & \SymExecutionFailureFlashAttentionCount & \SymExecutionFailureFlashAttentionRatio & \SymExecutionFailureTritonCount & \SymExecutionFailureTritonRatio & \SymExecutionFailureCutlassCount & \SymExecutionFailureCutlassRatio & \SymExecutionFailureDeepGemmCount & \SymExecutionFailureDeepGemmRatio & \SymExecutionFailureXformersCount & \SymExecutionFailureXformersRatio & \SymExecutionFailureBitsAndBytesCount & \SymExecutionFailureBitsAndBytesRatio \\
    \hline
    \SymFunctionalIncorrectness{} & \SymFunctionalIncorrectnessCount & \SymFunctionalIncorrectnessRatio & \SymFunctionalIncorrectnessFlashInferCount & \SymFunctionalIncorrectnessFlashInferRatio & \SymFunctionalIncorrectnessFlashAttentionCount & \SymFunctionalIncorrectnessFlashAttentionRatio & \SymFunctionalIncorrectnessTritonCount & \SymFunctionalIncorrectnessTritonRatio & \SymFunctionalIncorrectnessCutlassCount & \SymFunctionalIncorrectnessCutlassRatio & \SymFunctionalIncorrectnessDeepGemmCount & \SymFunctionalIncorrectnessDeepGemmRatio & \SymFunctionalIncorrectnessXformersCount & \SymFunctionalIncorrectnessXformersRatio & \SymFunctionalIncorrectnessBitsAndBytesCount & \SymFunctionalIncorrectnessBitsAndBytesRatio \\
    \hline
    \SymPerformanceDegradation{}  & \SymPerformanceDegradationCount & \SymPerformanceDegradationRatio & \SymPerformanceDegradationFlashInferCount & \SymPerformanceDegradationFlashInferRatio & \SymPerformanceDegradationFlashAttentionCount & \SymPerformanceDegradationFlashAttentionRatio & \SymPerformanceDegradationTritonCount & \SymPerformanceDegradationTritonRatio & \SymPerformanceDegradationCutlassCount & \SymPerformanceDegradationCutlassRatio & \SymPerformanceDegradationDeepGemmCount & \SymPerformanceDegradationDeepGemmRatio & \SymPerformanceDegradationXformersCount & \SymPerformanceDegradationXformersRatio & \SymPerformanceDegradationBitsAndBytesCount & \SymPerformanceDegradationBitsAndBytesRatio \\
    \hline
    \SymProgramHang{}             & \SymProgramHangCount & \SymProgramHangRatio & \SymProgramHangFlashInferCount & \SymProgramHangFlashInferRatio & \SymProgramHangFlashAttentionCount & \SymProgramHangFlashAttentionRatio & \SymProgramHangTritonCount & \SymProgramHangTritonRatio & \SymProgramHangCutlassCount & \SymProgramHangCutlassRatio & \SymProgramHangDeepGemmCount & \SymProgramHangDeepGemmRatio & \SymProgramHangXformersCount & \SymProgramHangXformersRatio & \SymProgramHangBitsAndBytesCount & \SymProgramHangBitsAndBytesRatio \\
    \hline
     Total                     & \FinalIssueCount & 100.00\% & \FlashInferFinalIssueCount & 12.33\% & \FlashAttentionFinalIssueCount & 9.33\% & \TritonFinalIssueCount & 25.67\% & \CutlassFinalIssueCount & 18.67\% & \DeepGemmFinalIssueCount & 1.33\% & \XformersFinalIssueCount & 6.33\% & \BitsAndBytesFinalIssueCount & 26.33\% \\
    \hline
  \end{tabular}%
  }
\end{table*}

From this pool, we finally constructed a repository-stratified sample of \FinalIssueCount issues, with each repository's sample size proportional to its number of candidate reports.
Each sampled issue was then examined alongside its discussion, diagnostic evidence, and available fix.
Reports containing only symptoms, speculative explanations, or competing explanations were excluded.
Each excluded report was replaced with an issue from the same repository whose creation time was closest to that of the excluded report.
This screening-and-replacement process continued until the final dataset contained \FinalIssueCount high-quality issues.
Table~\ref{tab:subjects} summarizes the raw collection, candidate dataset, and final sample for each repository.

\subsection{Data Labeling}

Symptom labeling focused on observable outcomes, drawing on each issue's title, the opening portion of its description, error messages, and reported behavior.
We adapted terminology and categories from prior empirical studies~\cite{Tan2014, dlbugch, Zhang2019, bugsinsidetf, Shen2021, Jia2021, Cao2022, Chen2023, Gao2023, Gao2025} to the context of GPU kernel libraries.
Each issue received one primary symptom label from ``{\SymBuildDeployment}''~\cite{Tan2014, dlbugch, Chen2023}, ``{\SymDeploymentFailure}''~\cite{chen2021empirical,Gao2023}, ``{\SymExecutionFailure}''~\cite{Tan2014, Shen2021, Gao2023, Gao2025}, ``{\SymFunctionalIncorrectness}''~\cite{Tan2014, dlbugch, Chen2023}, ``{\SymPerformanceDegradation}''~\cite{Tan2014, Jia2021}, or ``{\SymProgramHang}''~\cite{10.1145/3597204,Gao2025}.
Hardware, data type, trigger, downstream context, and diagnostic evidence were also recorded as auxiliary information.
Table~\ref{tab:taxonomy} presents the \SymptomCategoryCount{} resulting symptom categories.

Root-cause and fix analysis followed the resolution process of each issue, including its discussion, maintainer diagnosis, reproduction results, code analysis, and available repair.
Building on terminology from prior studies of deep-learning-related issues and bugs~\cite{Shen2021, dlbugch, Chen2023, Gao2023, Gao2025}, we adapted the categories according to the mechanisms observed in the selected reports.
Each issue was assigned a single root-cause category, for example, ``{\RcConcurrencyCoordination}''~\cite{Shen2021, Gao2025}, ``{\RcInterfaceMismatch}''~\cite{dlbugch, Chen2023, Gao2025}, ``{\RcNumericalSemantics}''~\cite{Shen2021}, or ``{\RcSoftwareEnvironment}''~\cite{Shen2021, Gao2025}.
Domain knowledge and patterns observed in the reports also guided our classification of root causes not covered by the existing taxonomy, including ``{\RcRuntimeDispatch}'' and ``{\RcCompilerTransformation}'',
which yielded further insights into GPU kernel library issues.
Finally, we organized the \RootCauseCategoryCount identified root-cause categories into \RootCauseDimensionCount dimensions to provide a comprehensive view.
Table~\ref{tab:rootpatterns} presents the overall taxonomy.

Two authors independently reviewed and labeled the final \ReviewSampleSize issues, considering their symptoms, root causes, and associated fixes.
Agreement was assessed at the issue level: a difference in either the symptom label or the root-cause label counted as a disagreement.
The resulting Cohen's kappa~\cite{Cohen1960} was \IndependentJointKappa, indicating a high level of labeling consistency.
We reconciled minor labeling disagreements through discussion and removed irrelevant issues outside the study's scope from the final dataset.

\section{What are the common symptoms?}\label{sec:symptoms}

The symptoms documented in the \FinalIssueCount{} issue reports fall into \SymptomCategoryCount{} categories.
Here, a symptom is the user-observed manifestation of an issue, usually described in the report's title or opening description.
Table~\ref{tab:taxonomy} summarizes the distribution of these categories across the seven studied libraries.

With \SymDeploymentFailureCount{} reports (\SymDeploymentFailureRatio{}), \emph{\SymDeploymentFailure{}} is the most frequently observed symptom.
This category covers failures during installation, dependency resolution, build orchestration, packaging, artifact selection, linking, loading, import, and startup.
In one \flashattention{} report, installation stopped while determining wheel build requirements because PyTorch was unavailable in the isolated build environment~\cite{flashattention-issue-258}.
Users worked around this problem by installing a compatible package version or disabling build isolation.

\emph{\SymBuildDeployment{}} accounts for \SymBuildDeploymentCount{} reports
(\SymBuildDeploymentRatio). Unlike deployment failures involving an existing
package, dependency, or artifact, these cases prevent source or generated code
from becoming an executable artifact. They include host, CUDA, DSL/JIT, IR,
PTX, SASS, and device-code compilation failures.
For example, compilation of a \flashattention{} forward kernel produced invalid Multi-Level Intermediate Representation (MLIR) with an SSA dominance violation after a \cutlass{} DSL upgrade~\cite{flashattention-issue-1918}.
Reverting the dependency avoided the failure.
Diagnosis of such problems requires minimized compiler reproducers and preserved intermediate representations, alongside regression tests across compiler and GPU targets.

\finding{1}{More than half of the observed symptoms arise before device execution: \SymDeploymentFailure{} (\SymDeploymentFailureCount{}; \SymDeploymentFailureRatio) and \SymBuildDeployment{} (\SymBuildDeploymentCount{}; \SymBuildDeploymentRatio).
Kernel-library reliability therefore depends on packages, dependencies, compilers, and generated artifacts as well as kernel code.}
{Testing should cover the full path from installation to kernel generation across supported combinations of frameworks, compilers, GPUs, and software-stack components.
Preflight diagnostics should identify incompatibilities before attempting to execute an operator.}

The \emph{\SymExecutionFailure{}} category contains \SymExecutionFailureCount{} reports (\SymExecutionFailureRatio).
Its scope extends from API validation and pre-launch dispatch failures to unsuccessful kernel launches and faults during device execution.
For example, a \flashinfer{} paged-KV prefill call failed before launch because no implementation supported a group size of three~\cite{flashinfer-issue-258}.
A \flashattention{} SM100 kernel, by contrast, reached launch but failed for small head dimensions because its stage selection exceeded shared-memory capacity~\cite{flashattention-issue-2591}.
These cases call for explicit operator contracts, dispatch-coverage tests, and launch and runtime diagnostics.

\emph{\SymProgramHang{}} accounts for \SymProgramHangCount{} reports (\SymProgramHangRatio{}).
These reports explicitly describe hangs or stuck processes, including coordination failures during concurrent execution and a lack of progress in kernels, builds, or compilation.
For example, a \cutlass{} Stream-K GEMM hung for several specific shapes because its separate-reduction path did not correctly coordinate the participating work~\cite{cutlass-issue-1801}.
Disabling this path allowed the same workload to complete.

Not all failures interrupt execution.
In \emph{\SymFunctionalIncorrectness{}} reports
(\SymFunctionalIncorrectnessCount{}; \SymFunctionalIncorrectnessRatio{}), an
operation completes but produces outputs or gradients that violate its intended
semantics.
One \flashattention{} report~\cite{flashattention-issue-2623} exposed a block-sparse attention path whose forward pass matched the sparse reference while its backward pass returned finite but severely inaccurate gradients.
The forward pass knew which attention blocks were enabled, but this information was not retained for the backward pass.
Backward therefore computed gradients as if dense attention had run.
Figure~\ref{fig:flashattention-sparse-backward} shows the repair: save the mask rule and backward block list during forward, then pass both to the backward kernel.
This restored agreement with an FP32 reference.
The case shows that a correct forward result does not guarantee correct training; tests must also compare gradients.

\finding{2}{\SymExecutionFailure{}, \SymProgramHang{}, and \SymFunctionalIncorrectness{} account for \SymExecutionFailureCount{} reports (\SymExecutionFailureRatio), \SymProgramHangCount{} reports (\SymProgramHangRatio), and \SymFunctionalIncorrectnessCount{} reports (\SymFunctionalIncorrectnessRatio), respectively.
  Together, these \ExecutionOutcomeIssueCount{} reports (\ExecutionOutcomeIssueRatio) capture three distinct outcomes: an operation may fail during invocation or execution, stop making progress, or complete with outputs that violate its intended semantics.}
{Tests should check both execution outcomes and result correctness, especially at boundary conditions.
  In particular, they should exercise resource-sensitive dimensions and compare outputs for semantic cases such as unequal sequence lengths.
  A successful kernel launch alone does not establish correct behavior.}

\begin{figure}[!t]
\begin{minted}[fontsize=\scriptsize,escapeinside=||,linenos]{python}
@staticmethod
def forward(...):
    out, lse = _flash_attn_fwd(
        q, k, v, mask_mod=mask_mod,
        block_sparse_tensors=block_sparse_tensors)
    # Save sparse state for backward.
    |\sout{\texttt{ctx.save\_for\_backward(q, k, v, out, lse)}}|
    |\textcolor{red}{\texttt{+ ctx.save\_for\_backward(q, k, v, out, lse, *(aux\_tensors or ()))}}|
    |\textcolor{red}{\texttt{+ ctx.mask\_mod = mask\_mod}}|
    |\textcolor{red}{\texttt{+ ctx.block\_sparse\_tensors\_bwd = block\_sparse\_tensors\_bwd}}|
    return out, lse

@staticmethod
def backward(ctx, dout, dlse):
    q, k, v, out, lse, *aux = ctx.saved_tensors
    dq, dk, dv = _flash_attn_bwd(
        q, k, v, out, dout, lse,
    # Restore sparse semantics in the gradient kernel.
    |\textcolor{red}{\texttt{+       mask\_mod=ctx.mask\_mod,}}|
        aux_tensors=aux if aux else None,
    |\textcolor{red}{\texttt{+       block\_sparse\_tensors=ctx.block\_sparse\_tensors\_bwd,}}|
        dlse=dlse)
    return dq, dk, dv, *((None,) * 30)
\end{minted}
\caption{An example of \SymFunctionalIncorrectness caused by losing block-sparse
state between forward and backward. The fix is to save the mask and backward
sparse metadata (lines 8--10) and pass them to the gradient kernel
(lines 19--21).}
\label{fig:flashattention-sparse-backward}
\end{figure}

\emph{\SymPerformanceDegradation{}} accounts for \SymPerformanceDegradationCount{} reports (\SymPerformanceDegradationRatio).
This category covers unexpected latency increases, reduced throughput, inefficient kernel execution, and tuning overhead.
For example, a \flashattention{} report documented a drop in training throughput on H100 GPUs from approximately 370 to 219 TFLOPS when the attention head dimension changed from 128 to 132~\cite{flashattention-issue-1141}.
The maintainer explained that the implementations favor tensor-core-aligned dimensions and may pad unconventional dimensions to multiples such as 32, 64, or 128.
The additional computation helps explain why a small change in head dimension can substantially reduce throughput.

\section{What are the common root causes and fixes?}\label{sec:rootcause}

We analyzed the root causes documented in the \FinalIssueCount{} issue reports.
Table~\ref{tab:rootpatterns} summarizes the resulting \RootCauseCategoryCount{} categories, organized into \RootCauseDimensionCount{} dimensions.

\begin{table*}[!t]
	\centering
	\caption{Classification of common root causes in the sample of \FinalIssueCount issues.}
	\label{tab:rootpatterns}
	\small
	\setlength{\tabcolsep}{5pt}
	\renewcommand{\arraystretch}{1.02}
	\begin{tabular}{|p{0.25\linewidth}|p{0.38\linewidth}|>{\centering\arraybackslash}p{0.07\linewidth}|>{\centering\arraybackslash}p{0.09\linewidth}|}
		\hline
		\bf Dimension & \bf Category & \bf No. & \bf Ratio \cr
		\hline
		\multirow{12}{=}{\RcRuntimeImplementation{}}
		 & \RcComputationalLogic{}      & \RcComputationalLogicCount & \RcComputationalLogicRatio \cr \cline{2-4}
		 & \RcNumericalSemantics{}      & \RcNumericalSemanticsCount & \RcNumericalSemanticsRatio \cr \cline{2-4}
		 & \RcIndexing{}                & \RcIndexingCount & \RcIndexingRatio \cr \cline{2-4}
		 & \RcDataLayout{}              & \RcDataLayoutCount & \RcDataLayoutRatio \cr \cline{2-4}
		 & \RcMemoryAccess{}            & \RcMemoryAccessCount & \RcMemoryAccessRatio \cr \cline{2-4}
		 & \RcRuntimeState{}            & \RcRuntimeStateCount & \RcRuntimeStateRatio \cr \cline{2-4}
		 & \RcConcurrencyCoordination{} & \RcConcurrencyCoordinationCount & \RcConcurrencyCoordinationRatio \cr \cline{2-4}
		 & \RcRuntimeDispatch{}         & \RcRuntimeDispatchCount & \RcRuntimeDispatchRatio \cr \cline{2-4}
		 & \RcResourcePlanning{}        & \RcResourcePlanningCount & \RcResourcePlanningRatio \cr \cline{2-4}
		 & \RcInefficientExecution{}    & \RcInefficientExecutionCount & \RcInefficientExecutionRatio \cr \cline{2-4}
		 & \RcInterfaceMismatch{}       & \RcInterfaceMismatchCount & \RcInterfaceMismatchRatio \cr \cline{2-4}
		 & Subtotal                 & \RcRuntimeImplementationCount & \RcRuntimeImplementationRatio \cr
		\hline
		\multirow{3}{=}{\RcCompilerInfrastructure{}}
		 & \RcCompilerTransformation{} & \RcCompilerTransformationCount & \RcCompilerTransformationRatio \cr \cline{2-4}
		 & \RcBuildConfiguration{}     & \RcBuildConfigurationCount & \RcBuildConfigurationRatio \cr \cline{2-4}
		 & Subtotal                & \RcCompilerInfrastructureCount & \RcCompilerInfrastructureRatio \cr
		\hline
		\multirow{3}{=}{\RcExecutionEnvironment{}}
		 & \RcHardwarePlatform{}    & \RcHardwarePlatformCount & \RcHardwarePlatformRatio \cr \cline{2-4}
		 & \RcSoftwareEnvironment{} & \RcSoftwareEnvironmentCount & \RcSoftwareEnvironmentRatio \cr \cline{2-4}
		 & Subtotal             & \RcExecutionEnvironmentCount & \RcExecutionEnvironmentRatio \cr
		\hline
		\multicolumn{2}{|l|}{Total} & \FinalIssueCount & 100.00\% \cr
		\hline
	\end{tabular}
\end{table*}

\subsection{\RcRuntimeImplementation{}-side Issues}

The \RcRuntimeImplementation{} dimension contains \RcRuntimeImplementationCount{} cases (\RcRuntimeImplementationRatio{}) across 11 categories.
These cases concern computation within the kernel library, runtime decisions, and interactions between integrated components.

\emph{\RcComputationalLogic{}} accounts for \RcComputationalLogicCount{} cases (\RcComputationalLogicRatio{}).
This category covers faulty predicates, branches, formulas, and special-case handling.
In one \deepgemm{} report, a TMA multicast heuristic used a Boolean indicating whether multicast targeted operand A as an array index~\cite{deepgemm-issue-190}.
After the preferred operand order was swapped, the heuristic checked operand B's legality but selected multicast on A.
This mismatch could select an invalid configuration.
The maintainer confirmed the logic error and corrected the array mapping so that the Boolean value and operand-legality entry used the same A/B order~\cite{deepgemm-fix-193}.
Repairs to this class of issues require locating the faulty control or computation logic in the kernel or its runtime selection code.

\emph{\RcNumericalSemantics{}} accounts for \RcNumericalSemanticsCount{} cases (\RcNumericalSemanticsRatio{}).
These cases concern numerical transformations and aggregations whose effective semantics differ from those intended or assumed.
The mechanisms include inappropriate precision or type conversion, quantization, scaling, clipping, finite-value substitution, rounding, contraction, arithmetic overflow, reduced-precision accumulation, and order-sensitive reduction.
For example, a \cutlass{} sparse GEMM with FP32 inputs produced values that differed substantially from a dense FP32 reference~\cite{cutlass-issue-1270}.
The maintainer explained that the sparse tensor-core path converted its inputs to TF32, whose shorter mantissa affected precision-sensitive values.
The reporter confirmed that the outputs agreed once the dense reference also used TF32 semantics.
Investigation therefore requires making the kernel's effective precision and accumulation mode explicit and comparing it with a reference that uses equivalent numerical semantics and appropriate tolerances.

\emph{\RcIndexing{}} covers \RcIndexingCount{} cases (\RcIndexingRatio{}).
Faulty coordinates, offsets, strides, identifiers, or index widths can select the wrong data or launch region.
An \xformers{} attention path narrowed the number of valid key columns in the current tile to a signed 16-bit parameter~\cite{xformers-issue-719}.
The tile loop supplied the remaining key count, while online softmax used that bound to include columns in both the row maximum and exponentiation.
Once the remaining span exceeded the signed 16-bit range, the predicates operated on a corrupted bound and attention returned NaN values.
As Figure~\ref{fig:xformers-valid-column-width} shows, the landed repair widened the parameter to \texttt{int}; the same commit added oversized query and key regression shapes~\cite{xformers-fix-719}.
In a downstream VAE decoding workflow, the affected outputs could produce black images.
Index widths must therefore cover the largest logical range before values enter tile-level predicates.

\begin{figure}[!t]
\begin{minted}[fontsize=\scriptsize,escapeinside=||,linenos]{cpp}
// Pass the remaining valid key columns to online softmax.
iterative_softmax(
    ..., p.num_keys - iter_key_start,
    iter_key_start == 0, ...);

CUTLASS_DEVICE static void iterative_softmax(
    ..., int8_t lane_id, int8_t thread_id, int8_t warp_id,
    |\sout{\texttt{int16\_t max\_col,}}|
    |\textcolor{red}{\texttt{+ int max\_col,}}|
    bool is_first, ...) {
    if (accum_n < max_col)
        max = cutlass::fast_max(max, frag[idx]);
    // ... row reduction omitted ...
    frag[idx] = accum_n < max_col ?
        exp2f(frag[idx] - mi_row) : accum_t(0.0);
}
\end{minted}
\caption{An example of Incorrect Indexing caused by storing the valid-column
bound in a signed 16-bit parameter (line 8), which overflows for long-context
attention. The fix is to widen the parameter to \texttt{int} (line 9).}
\Description{The unchanged call passes the remaining key count. The
struck-through parameter stores it as signed 16-bit, the red replacement uses
int, and unchanged predicates show where the bound controls online softmax.}
\label{fig:xformers-valid-column-width}
\vspace{-8pt}
\end{figure}

\emph{\RcDataLayout{}} accounts for \RcDataLayoutCount{} cases (\RcDataLayoutRatio{}).
These faults arise when tensor rank, stride, element width, or thread-to-data mapping does not match the representation expected by a kernel.
A \cutlass{} sparse GEMM kernel requires metadata in an interleaved layout named \texttt{Gemm::LayoutE}.
The calling example prepared row-major metadata but passed that original tensor to the kernel rather than the tensor produced by the required layout conversion~\cite{cutlass-issue-225}.
The kernel consequently interpreted the metadata using the wrong memory organization, and the correctness check observed an all-zero output.
The repair converted the metadata into a separate \texttt{Gemm::LayoutE} tensor and passed that tensor to the kernel.
Such cases call for validation of shape, stride, alignment, and layout contracts between caller-side data preparation and device execution.
Tests should include non-contiguous, transposed, and architecture-specific representations.

\emph{\RcMemoryAccess{}} comprises \RcMemoryAccessCount{} cases (\RcMemoryAccessRatio{}).
The surrounding layout may be correct, yet invalid pointers, bounds, masks, or aliasing relationships can still violate memory-access requirements.
For example, a \flashinfer{} sampling kernel~\cite{flashinfer-issue-384} supplied the same variable as both the input and output of \texttt{FlagHeads}, a parallel primitive in NVIDIA's CUB library~\cite{nvidiacub}.
The primitive requires separate storage for these arguments.
Aliasing them caused undefined behavior, and the sampler intermittently returned large out-of-vocabulary token identifiers instead of the only token with nonzero probability.
The repair introduced a separate output variable, after which the reporter confirmed that sampling returned the expected token.
Diagnosis requires checking pointer bounds, masks, alignment, and aliasing preconditions.
Memory sanitizers and small test inputs with unambiguous expected outputs can help expose these faults.

With only \RcRuntimeStateCount{} cases (\RcRuntimeStateRatio{}), \emph{\RcRuntimeState{}} is the smallest category in this dimension.
It concerns mutable state that is not correctly initialized, reset, migrated, or preserved across invocations.
In \bnb{}, \texttt{Linear8bitLt} and \texttt{MatMul8bitLt} retained \texttt{state.idx} across batches.
An index set for one batch could therefore remain active during the next.
The reporter observed worse evaluation results for Llama2-13B during training than in an independent inference run after training~\cite{bitsandbytes-issue-767}.
According to the reporter's diagnosis, a later batch reused index state from an earlier batch rather than its own current state, so matrix multiplication could select the wrong data and produce incorrect results.

\emph{\RcConcurrencyCoordination{}} accounts for \RcConcurrencyCoordinationCount{} cases (\RcConcurrencyCoordinationRatio{}).
These issues involve barriers, atomics, execution ordering, progress, and producer--consumer coordination.
One \deepgemm{} SM90 GEMM failed in the final two-CTA cluster~\cite{deepgemm-issue-147}.
For the reported shape, physical CTA B8 owned the ninth and last logical tile, whereas its launched peer B9 had no tile.
Each CTA-local empty barrier nevertheless required eight consumer-warp completion arrivals.
A faulty peer-liveness predicate treated B9 as active, so B8 routed four arrivals to its own barrier and four to B9's; B9 performed no consumer work and supplied no arrivals back to B8.
B8's barrier therefore remained at four of eight, and its TMA producer could not pass the normal wait before reusing that shared-memory stage.
As Figure~\ref{fig:deepgemm-peer-liveness} shows, the merged repair evaluates the alignment condition using \texttt{num\_n\_blocks}, detects the idle peer, and routes all eight arrivals locally~\cite{deepgemm-fix-149}.
Investigation of such issues requires checking barrier counts, participant liveness, and producer--consumer progress across all boundary tiles.

\begin{figure}[!t]
\begin{minted}{cpp}
// Each stage barrier needs 2 CTAs x 4 warps = 8 arrivals.
empty_barriers[i]->init(
    kNumTMAMulticast * kNumMathThreads / 32);

// Decide whether the paired multicast CTA will execute.
is_peer_cta_alive =
|\sout{\texttt{    kNum1DBlocksPerGroup \% kNumMulticast == 0 or}}|
|\textcolor{red}{\texttt{+   num\_n\_blocks \% kNumMulticast == 0 or}}|
    num_m_blocks % kNumMulticast == 0 or
    (next_block_idx ^ 1) < num_blocks;

// With multicast=2, lane 0/1 notify cluster CTA 0/1.
auto target_cta = scheduler.is_peer_cta_alive ?
    lane_idx : cute::block_rank_in_cluster();
lane_idx < kNumTMAMulticast ?
    empty_barriers[s]->arrive(target_cta) : void();
// ... consumer MMA and other pipeline work omitted ...
// B8's TMA producer waits locally before reusing stage s.
empty_barriers[s]->wait(
    (scheduler.current_iter * num_iterations + k_iter + 1) & 1);
\end{minted}
\caption{An example of a Concurrency Error caused by treating the idle peer CTA
B9 as active (line 7), which routes half of B8's releases to B9
(lines 13--16) and blocks B8's local wait (lines 19--20). The fix is to use the
runtime N-block count in the peer-liveness predicate (line 8).}
\Description{The struck-through predicate uses the compile-time number of
blocks per group. The red replacement uses the runtime number of N blocks. The
black code shows barrier initialization, arrival routing, and the local wait.}
\label{fig:deepgemm-peer-liveness}
\vspace{-8pt}
\end{figure}

\emph{\RcRuntimeDispatch{}} covers \RcRuntimeDispatchCount{} cases (\RcRuntimeDispatchRatio{}).
An otherwise supported operation may have no valid implementation selected because a kernel variant or dispatch rule is missing or overly restrictive.
In one \cutlass{} report, the operation selector rejected a mixed-FP8 GEMM whose operands A and B used different FP8 formats~\cite{cutlass-issue-1268}.
The underlying Hopper MMA implementation supported the combination, but a static assertion required identical operand types.
Template instantiation therefore stopped before the valid implementation could be selected.
The restriction was soon removed.

\emph{\RcResourcePlanning{}} accounts for \RcResourcePlanningCount{} cases (\RcResourcePlanningRatio{}).
These faults stem from incorrect calculations of storage, workspace, launch dimensions, or execution capacity.
A \cutlass{} Blackwell 2-SM GEMM example calculated its launch grid from the MMA tile size rather than the smaller tile assigned to each cooperative thread array (CTA)~\cite{cutlass-issue-2494}.
For a $1024 \times 1024$ output, the resulting grid was $4 \times 4$ instead of the required $8 \times 4$.
The launch created too few CTAs to cover all output tiles and left part of the GEMM result uncomputed.
The repair used the CTA tile shape to calculate the grid.
Validation should check both complete coverage of the output domain and whether shared memory, workspace, and worker requirements remain within hardware and runtime limits.
We assign a resource decision to this category when its calculation makes execution invalid or incomplete, rather than merely slower.

\emph{\RcInefficientExecution{}} comprises \RcInefficientExecutionCount{} cases (\RcInefficientExecutionRatio{}).
They arise from inefficient implementation choices that increase execution time or resource consumption despite producing the expected results.
Typical causes include redundant computation or recompilation, unnecessary synchronization and data transfers, excessive kernel launches, and uneven work distribution.
For example, a \triton{} attention-mask kernel~\cite{triton-issue-6480} used values that changed between invocations as \texttt{constexpr} parameters.
Each change triggered a new specialization, so JIT compilation and code generation took about 0.7~ms per call while the kernel itself ran in approximately 12~$\mu$s.
Repeated compilation blocked the CPU and created execution gaps in an SGLang serving workload.
A maintainer suggested checking whether compile-time arguments varied across calls, and the reporter confirmed that stabilizing them resolved the slowdown.
Diagnosis requires separating kernel execution time from compilation and framework overhead, then profiling the selected path for unnecessary specialization, transfers, synchronization, or other work.

\emph{\RcInterfaceMismatch{}} accounts for \RcInterfaceMismatchCount{} cases (\RcInterfaceMismatchRatio{}).
Following prior empirical work on software interface faults~\cite{Perry1985InterfaceFaults}, we use this category for problems at the boundaries between wrappers, frameworks, compiler APIs, runtimes, and kernels.
These problems include inconsistent types, shapes, layouts, symbols, metadata, mappings, and conversions, as well as violations of serialization, ownership, ordering, communication, or lifecycle requirements.
For example, a \flashattention{} change connected incompatible {\cutlass} pipeline parameter types~\cite{flashattention-issue-1476}.
A \triton{} benchmarking path invoked synchronization during CUDA Graph capture and violated the rules governing that interaction~\cite{triton-issue-10596}.
Addressing these faults requires explicit interface contracts and integration tests that check both data representations and complete interaction sequences.

\finding{3}{The \RcRuntimeImplementation{} cases span computation, state, dispatch, resource calculations, execution efficiency, and interfaces between integrated components.
Similar symptoms can therefore require different repairs.}
{Testing should combine numerical comparisons, memory and resource checks, interface-contract validation, and cross-component integration tests.}

\subsection{\RcCompilerInfrastructure{}-side Issues}

Compilers and build systems turn kernel implementations into executable device code.
This dimension contains \RcCompilerInfrastructureCount{} cases (\RcCompilerInfrastructureRatio{}) across the two categories in Table~\ref{tab:rootpatterns}.

\emph{\RcCompilerTransformation{}} is the larger category, with \RcCompilerTransformationCount{} cases (\RcCompilerTransformationRatio{}).
The faults occur in a library's built-in or directly integrated high-level DSL, JIT compiler, or code-generation pipeline.
Frontend, IR, optimization, or lowering transformations may fail during compilation or fail to preserve valid kernel semantics.
In \flashattention{}, a {\cutlass} DSL upgrade produced invalid MLIR with an SSA dominance violation~\cite{flashattention-issue-1918}.
An online-softmax fragment was used on a control-flow path where its definition did not dominate the use.
IR verification consequently stopped compilation before the SM90 forward kernel could be generated.
Reverting {\cutlass} DSL from 4.2.1 to 4.2.0 avoided the failure, and the maintainer later closed the report as fixed. 

\emph{\RcBuildConfiguration{}} accounts for \RcBuildConfigurationCount{} cases (\RcBuildConfigurationRatio{}).
This category concerns build scripts, targets, generators, dependency declarations, and compiler options that prevent construction or loading, or severely impede the build process.
For example, installation of a \flashattention{} CUDA extension appeared to freeze during compilation~\cite{flashattention-issue-215}.
Ninja~\cite{ninja} was not declared as a build requirement, so compilation fell back to a much slower non-parallel path rather than stopping with a clear missing-dependency message.
The maintainer added Ninja to the build requirements and restored the expected parallel compilation.

\finding{4}{Problems before device execution can arise from compiler transformations or build configuration.
The CUTLASS DSL case stopped at IR verification, whereas the missing Ninja requirement made compilation proceed slowly enough to appear stuck.}
{Distinguish compilation errors and genuine non-progress from slow build progress.
Preserve compiler artifacts and build logs, and verify build dependencies before concluding that compilation has stopped making progress.}

\subsection{\RcExecutionEnvironment{}-side Issues}

A kernel library needs a compatible hardware and software environment to build, load, and execute its specialized implementations.
The \RcExecutionEnvironmentCount{} cases (\RcExecutionEnvironmentRatio{}) in this dimension fall into two categories, as shown in Table~\ref{tab:rootpatterns}.

\emph{\RcHardwarePlatform{}} accounts for \RcHardwarePlatformCount{} cases (\RcHardwarePlatformRatio{}).
These issues arise when an operation requires a GPU architecture, compute capability, topology, instruction, or hardware resource that the available platform does not support.
For example, a \bnb{} user installed the CUDA extension successfully on an NVIDIA V100 but encountered an error when invoking INT8 matrix multiplication~\cite{bitsandbytes-issue-444}.
The maintainer explained that this implementation required INT8 Tensor Cores, which the V100 does not provide.
Changing software versions could not supply the missing hardware capability for that kernel path.
The user instead needed a compatible alternative implementation or newer hardware.
GPU capability checks before dispatch, supported fallbacks where available, and diagnostics that name the missing architectural feature can make these failures easier to understand and address.

\emph{\RcSoftwareEnvironment{}} is the largest individual root-cause category, with \RcSoftwareEnvironmentCount{} cases (\RcSoftwareEnvironmentRatio{}).
It covers incompatibilities among the library's dependencies and execution environment, including GPU drivers, CUDA/ROCm runtimes, deep learning frameworks, operating systems, ABIs, and their version requirements.
In one \flashattention{} report, a call failed at runtime although CUDA 11.6 was available on the host~\cite{flashattention-issue-217}.
The installed PyTorch package had been compiled against CUDA 11.3, below the library's requirement.
The same input worked after the reporter installed a PyTorch build compiled for CUDA 11.6.
This comparison isolated the failure to the software combination rather than the GPU or attention kernel.
Diagnosis should record both installed and build-time versions, check compatibility across the complete stack, and reproduce the problem with a documented supported combination.

\finding{5}{The \RcExecutionEnvironment{} dimension accounts for \RcExecutionEnvironmentCount{} reports (\RcExecutionEnvironmentRatio{}).
\RcSoftwareEnvironment{} is the largest individual category, with \RcSoftwareEnvironmentCount{} reports (\RcSoftwareEnvironmentRatio{}).}
{Compatibility checks should consider package, framework, compiler, CUDA, and driver versions together with the GPU architecture.
Updating a single component does not establish that the resulting combination is supported.}

\section{Current testing and debugging practices}\label{sec:evidence}

The selected reports document not only failures and fixes but also the tests and diagnostic steps used to investigate them.
We summarize these accounts to examine how developers investigate and resolve GPU kernel library issues.
Our evidence is limited to what appears in the reports and their discussions; it does not establish whether developers also used practices that went unrecorded.

\subsection{Testing Practices}

GPU kernel tests must check both the computed results and the implementation's behavior under supported execution conditions.
Drawing on the reported cases and guidance from NVIDIA and PyTorch~\cite{nvidiacudabestpractices,nvidiacomputesanitizer,pytorchbenchmark}, we summarize five complementary testing practices.

\emph{Compatibility and Integration Testing.}
Developers record the hardware and software stack and use representative supported inputs to check interactions among frameworks, wrappers, compilers, and kernels.
In one \cutlass{} report, a Turing Tensor Core test failed with CUDA 10.1 because the operation required CUDA 10.2 or later~\cite{cutlass-issue-80}.
Such checks can expose platform, version, and interface incompatibilities before they surface as opaque kernel errors.

\emph{Differential Testing.}
The optimized kernel and a simpler reference receive the same input, and their outputs are compared under consistent settings and an appropriate tolerance.
The reference itself also requires scrutiny.
In one \flashattention{} report, gradients disagreed with an FP16 reference but matched the same calculation in FP32~\cite{flashattention-issue-385}.
The discrepancy arose from rounding error in the reference rather than a defect in the optimized kernel.
Differential testing can reveal silent incorrectness, but further diagnosis is needed to establish the source of a mismatch.

\emph{Boundary Value Analysis.}
Developers vary one dimension, sequence-length relationship, alignment, or layout at a time while keeping other settings fixed.
In a \flashattention{} issue, a sweep showed failures at head dimensions 8 and 16 but not 32, which helped isolate an incorrect shared-memory calculation~\cite{flashattention-issue-2591}.
A separate report required unequal query and key lengths to reveal a masking error~\cite{flashattention-issue-282}.
Tests restricted to common shapes can leave such faults undetected.

\emph{Dynamic Analysis.}
NVIDIA Compute Sanitizer~\cite{nvidiacomputesanitizer} provides checks for memory access, races, initialization, and synchronization.
In a \flashinfer report involving a 48-rank CuTe DSL~\cite{nvidiacutedsl} GEMM, it detected an illegal memory access~\cite{flashinfer-issue-1618}.

\emph{Performance Regression Testing.}
Developers compare a fixed workload against a baseline after warm-up and measure framework, compilation, and kernel time separately~\cite{pytorchbenchmark}.
A \triton{} report illustrates the importance of this separation: each invocation spent about 0.7~ms in repeated JIT compilation, while the kernel itself took approximately 12~$\mu$s~\cite{triton-issue-6480}.
These measurements placed the slowdown outside GPU execution.
NVIDIA Nsight Systems~\cite{nvidiansightsystems} and Nsight Compute~\cite{nvidiansightcompute} can further help identify the critical path and kernel bottlenecks.

\subsection{Debugging Practices}

Diagnosis typically begins with the evidence attached to a report: error messages, stack traces, generated compiler output, and details of the hardware and software environment.
Developers then attempt to reproduce the failure with the supplied input and inspect the relevant wrapper, compiler, or kernel code.
For straightforward cases, the remedy may be a code correction, a configuration change, or a switch to a compatible software version.

More difficult cases require a small reproducer that retains the conditions responsible for the failure.
A compact \flashinfer{} sampling script, for example, returned invalid token identifiers, but the maintainer initially could not reproduce the behavior.
A comparison of CUDA environments revealed that only the failing environment selected a CUB~\cite{nvidiacub} operation subject to an aliasing restriction.
Locating the fault therefore required both the reproducing input and the relevant software version.

Specialized diagnostics help narrow the investigation to a particular layer.
NVIDIA Compute Sanitizer can identify memory and synchronization faults, while compiler intermediate representation (IR) and version comparisons provide evidence for failures before execution.
In one \flashattention{} report, an IR verification error appeared after a CUTLASS DSL upgrade and disappeared when the dependency was reverted.
The generated IR localized the failure to an SSA dominance violation and supplied evidence for an upstream report.
The issue did not, however, identify the exact repair commit.

Some failures remain difficult to reproduce because they depend on a particular shape, layout, backend, hardware target, or software combination.
Developers investigate these dependencies by changing one factor at a time, rerunning the reproducer, and comparing outcomes.
An issue first observed in a model or serving system may require investigation across several library boundaries.
A proposed fix should be checked against the original reproducer and its expected result.
The disappearance of an error after a workload change does not, by itself, establish that the underlying fault has been repaired.

\section{Discussion}\label{sec:discussion}

\subsection{Threats to Validity}\label{sec:threats}

\emph{Threats to Internal Validity}.
Classifying symptoms and root causes and interpreting fixes require judgment, particularly when an issue is complex or its supporting evidence is incomplete.
Issue descriptions, discussions, and linked fixes inform these judgments but do not eliminate subjectivity.
To limit this threat, we established explicit coding rules and retained the evidence supporting each classification.
Two reviewers independently examined all \ReviewSampleSize{} cases.
Their agreement, measured with Cohen's kappa~\cite{Cohen1960}, was \IndependentJointKappa{}, indicating a high level of consistency.
We further discussed disagreements collectively and sought consensus before making critical classification decisions.

\emph{Threats to External Validity}.
The selection of libraries and issue reports limits how broadly our findings can be generalized.
Our subjects comprise seven open-source GPU kernel libraries that cover attention, GPU programming, GEMM, inference, and quantization.
They include both research-originated projects and reusable industrial libraries.
Their use extends across training ecosystems such as Megatron-LM, PyTorch, Transformers, and DeepSpeed, as well as serving systems like vLLM and SGLang.
These connections support the relevance of the observed issues beyond a single repository.
They do not, however, establish that the findings apply equally to other GPU kernel projects, proprietary kernels, or complete training and serving systems.
In addition, replacing reports without identifiable causes may favor well-documented cases.
To mitigate this threat, we therefore distinguish repository-specific observations from aggregate results and avoid generalizing findings that may be specific to the seven studied libraries to GPU kernel libraries as a whole.

\subsection{Generality of Our Study}\label{sec:generality}

Our findings can be relevant to other GPU kernel projects that share the technical characteristics of the seven studied libraries.
The studied libraries connect Python or C++ interfaces with CUDA, Triton, or generated device code through similar heterogeneous programming stacks.
They also implement common LLM operators---attention, GEMM, quantization, and low-precision computation---with related constraints on tensor shapes, layouts, numerical precision, dispatch, and hardware resources.
Their deployment places them within comparable framework, compiler, driver, and GPU environments, with integrations into both training and serving systems.
These commonalities suggest that similar failure mechanisms may recur elsewhere and that the testing and debugging lessons drawn from our cases may transfer to other projects.

Evidence from related systems provides additional support for this interpretation.
Empirical studies of deep learning software have documented problems involving configuration, APIs, dependencies, and execution environments~\cite{Zhang2018, Zhang2019, bugsinsidetf, Gao2023,Humbatova2020DLFaults}.
Studies of deep learning compilers likewise report defects in optimization and code generation~\cite{Shen2021}.
GPU verification and testing research addresses races, barrier divergence, memory-access errors, and compiler defects~\cite{Betts2012GPUVerify, Li2012GKLEE, Peng2018CURD, Jiang2020CUDAsmith, Chakraborty2025GPUMC}.
The overlap between these independent lines of work and our taxonomy provides further evidence that the identified mechanisms are not unique to the seven repositories examined here.

\subsection{Future Research Directions}\label{sec:research}

The findings suggest several opportunities to improve tool support and library design.
The following directions could help developers detect incompatibilities earlier, diagnose failures more precisely, and reproduce problems reliably.

\noindent\textbf{Tool Support:}

\emph{Compatibility Checker}.
\RcExecutionEnvironment{} accounts for \RcExecutionEnvironmentCount{} cases (\RcExecutionEnvironmentRatio), while \SymDeploymentFailure{} is the largest symptom category, with \SymDeploymentFailureCount{} cases (\SymDeploymentFailureRatio).
These findings motivate a compatibility checker that examines the GPU, driver, runtime, compiler, framework, package, and binary ABI before a build or execution begins.
The checker should distinguish installed software versions from those used to build frameworks and binary extensions, since these may differ.
When a requirement is violated, it could identify the conflicting components, explain the incompatibility, and recommend a supported combination.
Such diagnostics would give developers a concrete starting point instead of leaving them to interpret a generic build or GPU kernel error.

\emph{Numerical Debugging Tools}.
\SymFunctionalIncorrectness{} accounts for \SymFunctionalIncorrectnessCount{} cases (\SymFunctionalIncorrectnessRatio), with causes that include \RcComputationalLogic{}, \RcNumericalSemantics{}, \RcIndexing{}, \RcDataLayout{}, and \RcMemoryAccess{}.
Numerical debugging tools could compare kernel and reference values at intermediate stages and track the associated shapes, data types, accumulation modes, masks, and tolerances.
TensorFuzz~\cite{Odena2019TensorFuzz} and Herbie~\cite{Panchekha2015Herbie} illustrate techniques for detecting anomalous numerical behavior and improving floating-point expressions.
Adapting such techniques to specialized GPU kernels requires careful treatment of their numerical contracts.

\emph{Compiler Debugging Tools}.
\RcCompilerTransformation{} accounts for \RcCompilerTransformationCount{} cases (\RcCompilerTransformationRatio).
A compiler debugging tool could reduce the triggering kernel and preserve its source, generated IR, compiler version, compilation options, and target architecture for differential and regression analysis.
IR snapshots could help identify the transformation at which compilation fails or valid kernel semantics are lost.
Compiler-testing systems such as CUDAsmith~\cite{Jiang2020CUDAsmith} and NNSmith~\cite{Liu2023NNSmith} provide techniques for generating programs that expose compiler defects.
Combining test generation with failure reduction and artifact preservation could make reports easier to investigate and turn confirmed defects into focused regression tests.

\noindent\textbf{Library Enhancements:}

\emph{Contract-aware Validation}.
Pre-launch API failures and issues involving \RcRuntimeDispatch{}, \RcDataLayout{}, \RcMemoryAccess{}, and \RcResourcePlanning{} often reflect implicit kernel requirements that are violated or calculated incorrectly.
Libraries could express contracts for shapes, data types, layouts, alignment, aliasing, supported in-place operations, and shared-memory requirements as executable checks in their wrappers and dispatch layers.
These checks should run before kernel launch and report both the expected and supplied values.
Where a requirement depends on the selected implementation or GPU architecture, the diagnostic should identify that dependency explicitly.

\emph{Record and Replay}.
Many kernel failures depend on a particular input, selected implementation, compiler artifact, software stack, or GPU architecture.
Reconstructing this combination from an issue discussion can be difficult, especially after dependencies or implementations have changed.
Libraries could record the operation, tensor metadata, dispatch decision, software versions, target architecture, and generated IR needed to replay the failure~\cite{10.5555/1855741.1855755}.
Metadata alone may be insufficient for value-dependent failures, so replay support should also preserve the triggering input or a reproducible procedure for generating it.
Once the cause is confirmed, the recorded case could become a regression test that recreates the relevant configuration and checks the expected result or behavior.

\section{Related Work}\label{sec:related}

Attention, GEMM, and quantization recur across large-scale LLM workloads, making specialized GPU kernels valuable to many models and systems.
TVM~\cite{tvm}, Ansor~\cite{Zheng2020Ansor}, Triton~\cite{Tillet2019Triton}, and FlashAttention~\cite{Dao2022FlashAttention} reduce memory traffic and improve throughput through techniques such as fusion, tiling, low-precision computation, and hardware-specific instructions.
These optimizations also impose complex requirements on compilation, data layout, dispatch, and compatibility.

Empirical research has examined defects at several levels of the AI software stack.
Studies of TensorFlow and other deep learning frameworks document API, configuration, data, and numerical problems, as well as silent behavioral bugs~\cite{Zhang2018,Zhang2019,bugsinsidetf,Tambon2021SilentBI,Humbatova2020DLFaults,dlbugch}.
Other work addresses performance bugs in TensorFlow and PyTorch~\cite{9978257}, deployment faults in deep learning applications~\cite{chen2021empirical}, production job failures~\cite{10.1145/3377811.3380362}, and distributed-training challenges~\cite{10.1145/3597204}.
Gao et al.~\cite{Gao2025} analyzed issues in DeepSpeed, Megatron-LM, and Transformers, with attention to their symptoms, root causes, fixes, and testing and debugging practices.
Shen et al.~\cite{Shen2021} investigated defects in deep learning compilers.

Complementary research on GPU verification and compiler testing targets races, barrier divergence, memory faults, and compiler defects through tools such as GPUVerify~\cite{Betts2012GPUVerify}, GKLEE~\cite{Li2012GKLEE}, CURD~\cite{Peng2018CURD}, GPUMC~\cite{Chakraborty2025GPUMC}, CUDAsmith~\cite{Jiang2020CUDAsmith}, and NNSmith~\cite{Liu2023NNSmith}.
TensorFuzz~\cite{Odena2019TensorFuzz} and Herbie~\cite{Panchekha2015Herbie} address anomalous floating-point behavior and numerical accuracy.

The empirical literature largely concerns applications, frameworks, training platforms, and compilers, while existing testing techniques focus on particular fault classes.
Our study instead examines reported symptoms, root causes, fixes, and engineering practices across the lifecycle of seven GPU kernel libraries.
The analysis spans the \RcRuntimeImplementation{}, \RcCompilerInfrastructure{}, and \RcExecutionEnvironment{} dimensions.

\section{Conclusion}\label{sec:conclusion}

Our empirical analysis of \FinalIssueCount issues from seven GPU kernel libraries identified \SymptomCategoryCount{} symptom categories and \RootCauseCategoryCount{} root-cause categories across \RootCauseDimensionCount{} dimensions.
We also documented typical fixes and reported testing and debugging practices, and outlined research directions for tool support and library enhancements.
These findings provide a clearer understanding of GPU kernel library issues and practical guidance for developing more reliable and efficient LLM systems.

\clearpage

\bibliographystyle{IEEEtran}
\bibliography{main}

@article{Jia2021,
title = {The symptoms, causes, and repairs of bugs inside a deep learning library},
journal = {Journal of Systems and Software},
volume = {177},
pages = {110935},
year = {2021},
issn = {0164-1212},
doi = {https://doi.org/10.1016/j.jss.2021.110935},
url = {https://www.sciencedirect.com/science/article/pii/S0164121221000327},
author = {Li Jia and Hao Zhong and Xiaoyin Wang and Linpeng Huang and Xuansheng Lu}
}

@article{Tan2014,
author = {Tan, Lin and Liu, Chen and Li, Zhenmin and Wang, Xuanhui and Zhou, Yuanyuan and Zhai, Chengxiang},
title = {Bug characteristics in open source software},
year = {2014},
issue_date = {December  2014},
publisher = {Kluwer Academic Publishers},
address = {USA},
volume = {19},
number = {6},
issn = {1382-3256},
url = {https://doi.org/10.1007/s10664-013-9258-8},
doi = {10.1007/s10664-013-9258-8},
journal = {Empirical Softw. Engg.},
month = dec,
pages = {1665–1705},
numpages = {41}
}

@inproceedings{Cao2022,
author = {Cao, Junming and Chen, Bihuan and Sun, Chao and Hu, Longjie and Wu, Shuaihong and Peng, Xin},
title = {Understanding performance problems in deep learning systems},
year = {2022},
isbn = {9781450394130},
publisher = {Association for Computing Machinery},
address = {New York, NY, USA},
url = {https://doi.org/10.1145/3540250.3549123},
doi = {10.1145/3540250.3549123},
booktitle = {Proceedings of the 30th ACM Joint European Software Engineering Conference and Symposium on the Foundations of Software Engineering},
pages = {357–369},
numpages = {13},
location = {Singapore, Singapore},
series = {ESEC/FSE 2022}
}

@inproceedings{Gao2025,
  author = {Gao, Yanjie and Lu, Ruiming and Lin, Haoxiang and Chen, Yueguo},
  title = {An Empirical Study of Issues in Large Language Model Training Systems},
  booktitle = {Proceedings of the 33rd ACM International Conference on the Foundations of Software Engineering},
  series = {FSE Companion '25},
  year = {2025},
  pages = {122--133},
  publisher = {Association for Computing Machinery},
  doi = {10.1145/3696630.3728538}
}

@misc{nvidiacub,
  author = {{NVIDIA Corporation}},
  title = {{CUB}: Cooperative Primitives for {CUDA} {C++}},
  year = {2024},
  howpublished = {\url{https://github.com/nvidia/cub}}
}

@misc{nvidiacudabestpractices,
  author = {{NVIDIA Corporation}},
  title = {{CUDA C++ Best Practices Guide}},
  year = {2026},
  howpublished = {\url{https://docs.nvidia.com/cuda/cuda-c-best-practices-guide/}},
  note = {Accessed: September 17, 2026}
}

@misc{nvidiacomputesanitizer,
  author = {{NVIDIA Corporation}},
  title = {{Compute Sanitizer User Manual}},
  year = {2026},
  howpublished = {\url{https://docs.nvidia.com/compute-sanitizer/ComputeSanitizer/}},
  note = {Accessed: September 17, 2026}
}

@misc{nvidiansightsystems,
  author = {{NVIDIA Corporation}},
  title = {{NVIDIA Nsight Systems User Guide}},
  year = {2026},
  howpublished = {\url{https://docs.nvidia.com/nsight-systems/UserGuide/}},
  note = {Accessed: September 17, 2026}
}

@misc{nvidiansightcompute,
  author = {{NVIDIA Corporation}},
  title = {{NVIDIA Nsight Compute User Guide}},
  year = {2026},
  howpublished = {\url{https://docs.nvidia.com/nsight-compute/NsightCompute/}},
  note = {Accessed: September 17, 2026}
}

@misc{pytorchbenchmark,
  author = {{PyTorch Contributors}},
  title = {{PyTorch Benchmark}},
  year = {2025},
  howpublished = {\url{https://docs.pytorch.org/tutorials/recipes/recipes/benchmark.html}},
  note = {Accessed: September 17, 2026}
}

@article{Cohen1960,
  author  = {Cohen, Jacob},
  title   = {A Coefficient of Agreement for Nominal Scales},
  journal = {Educational and Psychological Measurement},
  year    = {1960},
  volume  = {20},
  number  = {1},
  pages   = {37--46},
  doi     = {10.1177/001316446002000104}
}

@inproceedings{Zhang2018,
author = {Zhang, Yuhao and Chen, Yifan and Cheung, Shing-Chi and Xiong, Yingfei and Zhang, Lu},
title = {An empirical study on {TensorFlow} program bugs},
year = {2018},
isbn = {9781450356992},
publisher = {Association for Computing Machinery},
address = {New York, NY, USA},
url = {https://doi.org/10.1145/3213846.3213866},
doi = {10.1145/3213846.3213866},
booktitle = {Proceedings of the 27th ACM SIGSOFT International Symposium on Software Testing and Analysis},
pages = {129–140},
numpages = {12},
location = {Amsterdam, Netherlands},
series = {ISSTA 2018}
}

@inproceedings{Zhang2019,
author={Zhang, Tianyi and Gao, Cuiyun and Ma, Lei and Lyu, Michael and Kim, Miryung},
booktitle={2019 IEEE 30th International Symposium on Software Reliability Engineering (ISSRE)},
title={An Empirical Study of Common Challenges in Developing Deep Learning Applications},
year={2019},
volume={},
number={},
pages={104-115},
doi={10.1109/ISSRE.2019.00020}
}

@inproceedings{Shen2021,
  author = {Shen, Q. and Ma, H. and Chen, J. and Tian, Y. and Cheung, S.-C. and Chen, X.},
  title = {A Comprehensive Study of Deep Learning Compiler Bugs},
  booktitle = {Proceedings of the 29th ACM Joint Meeting on European Software Engineering Conference and Symposium on the Foundations of Software Engineering},
  year = {2021},
  pages = {968--980},
}

@inproceedings{pytorch,
 author = {Paszke, Adam and Gross, Sam and Massa, Francisco and Lerer, Adam and Bradbury, James and Chanan, Gregory and Killeen, Trevor and Lin, Zeming and Gimelshein, Natalia and Antiga, Luca and Desmaison, Alban and Kopf, Andreas and Yang, Edward and DeVito, Zachary and Raison, Martin and Tejani, Alykhan and Chilamkurthy, Sasank and Steiner, Benoit and Fang, Lu and Bai, Junjie and Chintala, Soumith},
 booktitle = {Advances in Neural Information Processing Systems},
 editor = {H. Wallach and H. Larochelle and A. Beygelzimer and F. d\textquotesingle Alch\'{e}-Buc and E. Fox and R. Garnett},
 pages = {},
 publisher = {Curran Associates, Inc.},
 title = {{PyTorch}: An Imperative Style, High-Performance Deep Learning Library},
 url = {https://proceedings.neurips.cc/paper_files/paper/2019/file/bdbca288fee7f92f2bfa9f7012727740-Paper.pdf},
 volume = {32},
 year = {2019}
}

@inproceedings {tvm,
author = {Tianqi Chen and Thierry Moreau and Ziheng Jiang and Lianmin Zheng and Eddie Yan and Haichen Shen and Meghan Cowan and Leyuan Wang and Yuwei Hu and Luis Ceze and Carlos Guestrin and Arvind Krishnamurthy},
title = {{TVM}: An Automated {End-to-End} Optimizing Compiler for Deep Learning},
booktitle = {13th USENIX Symposium on Operating Systems Design and Implementation (OSDI 18)},
year = {2018},
isbn = {978-1-939133-08-3},
address = {Carlsbad, CA},
pages = {578--594},
url = {https://www.usenix.org/conference/osdi18/presentation/chen},
publisher = {USENIX Association},
month = oct,
}

@inproceedings{dlbugch,
  author    = {Md Johirul Islam and Giang Nguyen and Rangeet Pan and Hridesh Rajan},
  title     = {A Comprehensive Study on Deep Learning Bug Characteristics},
  booktitle = {Proceedings of the 27th ACM Joint Meeting on ESEC/FSE},
  year      = {2019},
  pages     = {510--520},
}

@inproceedings{bugsinsidetf,
  author = {Jia, L. and Zhong, H. and Wang, X. and Huang, L. and Lu, X.},
  title = {An Empirical Study on Bugs Inside TensorFlow},
  booktitle = {Database Systems for Advanced Applications: 25th International Conference, DASFAA 2020},
  year = {2020},
  pages = {604--620}
}

@article{Tambon2021SilentBI,
  title={Silent Bugs in Deep Learning Frameworks: An Empirical Study of Keras and TensorFlow},
  author={Florian Tambon and Amin Nikanjam and Le An and Foutse Khomh and Giuliano Antoniol},
  journal={ArXiv},
  year={2021},
  volume={abs/2112.13314}
}

@INPROCEEDINGS{9978257,
  author={Makkouk, Tarek and Kim, Dong Jae and Chen, Tse-Hsun Peter},
  booktitle={2022 IEEE International Conference on Software Maintenance and Evolution (ICSME)}, 
  title={An Empirical Study on Performance Bugs in Deep Learning Frameworks}, 
  year={2022},
  volume={},
  number={},
  pages={35-46},
  }

@inproceedings{10.1145/3377811.3380362,
author = {Zhang, Ru and Xiao, Wencong and Zhang, Hongyu and Liu, Yu and Lin, Haoxiang and Yang, Mao},
title = {An Empirical Study on Program Failures of Deep Learning Jobs},
year = {2020},
isbn = {9781450371216},
publisher = {Association for Computing Machinery},
address = {New York, NY, USA},
url = {https://doi.org/10.1145/3377811.3380362},
doi = {10.1145/3377811.3380362},
booktitle = {Proceedings of the ACM/IEEE 42nd International Conference on Software Engineering},
pages = {1159–1170},
numpages = {12},
location = {Seoul, South Korea},
series = {ICSE '20}
}

@article{10.1145/3597204,
author = {Liu, Xuanzhe and Gu, Diandian and Chen, Zhenpeng and Wen, Jinfeng and Zhang, Zili and Ma, Yun and Wang, Haoyu and Jin, Xin},
title = {Rise of Distributed Deep Learning Training in the Big Model Era: From a Software Engineering Perspective},
year = {2023},
issue_date = {November 2023},
publisher = {Association for Computing Machinery},
address = {New York, NY, USA},
volume = {32},
number = {6},
issn = {1049-331X},
url = {https://doi.org/10.1145/3597204},
doi = {10.1145/3597204},
journal = {ACM Trans. Softw. Eng. Methodol.},
month = sep,
articleno = {156},
numpages = {26}
}

@article{shoeybi2019megatron,
  author       = {Mohammad Shoeybi and
                  Mostofa Patwary and
                  Raul Puri and
                  Patrick LeGresley and
                  Jared Casper and
                  Bryan Catanzaro},
  title        = {{Megatron-LM}: Training Multi-Billion Parameter Language Models Using
                  Model Parallelism},
  journal      = {CoRR},
  volume       = {abs/1909.08053},
  year         = {2019},
  url          = {https://doi.org/10.48550/arXiv.1909.08053},
  doi          = {10.48550/ARXIV.1909.08053},
  eprinttype   = {arXiv},
  eprint       = {1909.08053},
  bibsource    = {dblp computer science bibliography, https://dblp.org}
}

@INPROCEEDINGS{Gao2023,
  author={Gao, Yanjie and Shi, Xiaoxiang and Lin, Haoxiang and Zhang, Hongyu and Wu, Hao and Li, Rui and Yang, Mao},
  booktitle={2023 IEEE/ACM 45th International Conference on Software Engineering: Software Engineering in Practice (ICSE-SEIP)}, 
  title={An Empirical Study on Quality Issues of Deep Learning Platform}, 
  year={2023},
  volume={},
  number={},
  pages={455-466},
  }

@inproceedings{10.5555/1855741.1855755,
author = {Guo, Zhenyu and Wang, Xi and Tang, Jian and Liu, Xuezheng and Xu, Zhilei and Wu, Ming and Kaashoek, M. Frans and Zhang, Zheng},
title = {R2: an application-level kernel for record and replay},
year = {2008},
publisher = {USENIX Association},
address = {USA},
booktitle = {Proceedings of the 8th USENIX Conference on Operating Systems Design and Implementation},
pages = {193--208},
numpages = {16},
location = {San Diego, California},
series = {OSDI '08}
}

@INPROCEEDINGS{8115662,
  author={Di Franco, Anthony and Guo, Hui and Rubio-González, Cindy},
  booktitle={2017 32nd IEEE/ACM International Conference on Automated Software Engineering (ASE)}, 
  title={A comprehensive study of real-world numerical bug characteristics}, 
  year={2017},
  volume={},
  number={},
  pages={509-519},
  }

@misc{githubrestapi,
  author       = {{GitHub}},
  title        = {GitHub REST API Documentation},
  howpublished = {\url{https://docs.github.com/en/rest}},
  note         = {Accessed September 15, 2026}
}

@misc{transformersgit,
    author = {{Hugging Face}},
    title = {Transformers},
    howpublished = {\url{https://github.com/huggingface/transformers}},
    year = {2024}
}

@inproceedings{chen2021empirical,
  title={An empirical study on deployment faults of deep learning based mobile applications},
  author={Chen, Zhenpeng and Yao, Huihan and Lou, Yiling and Cao, Yanbin and Liu, Yuanqiang and Wang, Haoyu and Liu, Xuanzhe},
  booktitle={2021 IEEE/ACM 43rd International Conference on Software Engineering (ICSE)},
  pages={674--685},
  year={2021},
  organization={IEEE}
}

@article{Chen2023,
author = {Chen, Junjie and Liang, Yihua and Shen, Qingchao and Jiang, Jiajun and Li, Shuochuan},
title = {Toward Understanding Deep Learning Framework Bugs},
year = {2023},
issue_date = {November 2023},
publisher = {Association for Computing Machinery},
address = {New York, NY, USA},
volume = {32},
number = {6},
issn = {1049-331X},
url = {https://doi.org/10.1145/3587155},
doi = {10.1145/3587155},
journal = {ACM Trans. Softw. Eng. Methodol.},
month = sep,
articleno = {135},
numpages = {31}
}

@misc{ninja,
  author = {Ninja},
  title = {Ninja},
  year = {2011},
  howpublished = {\url{https://ninja-build.org}},
  note = {Ninja is a small build system with a focus on speed.}
}

@misc{flashinfer-issue-384,
  author = {FlashInfer},
  title = {{top\_k\_top\_p\_sampling\_from\_probs} occasionally outputs invalid token IDs},
  year = {2024},
  url = {https://github.com/flashinfer-ai/flashinfer/issues/384},
}

@misc{flashinfer-issue-258,
  author = {FlashInfer},
  title = {[Bug report] BatchPrefillWithPagedKVCachePyTorchWrapper failed to dispatch group\_size 3},
  year = {2024},
  url = {https://github.com/flashinfer-ai/flashinfer/issues/258},
}

@misc{flashinfer-issue-1618,
  author = {FlashInfer},
  title = {CuTeDSL GEMM illegal access on 48 ranks},
  year = {2025},
  url = {https://github.com/flashinfer-ai/flashinfer/issues/1618},
}

@misc{flashattention-issue-258,
  author = {FlashAttention},
  title = {Error: Install pip install flash-attn},
  year = {2023},
  url = {https://github.com/Dao-AILab/flash-attention/issues/258},
}

@misc{flashattention-issue-1918,
  author = {FlashAttention},
  title = {Cute-DSL SM90 fwd produces error after bumping to nvidia-cutlass-dsl-4.2.1},
  year = {2025},
  url = {https://github.com/Dao-AILab/flash-attention/issues/1918},
}

@misc{flashattention-issue-2591,
  author = {FlashAttention},
  title = {[FA4, CuTe, SM100] \_flash\_attn\_fwd fails at launch for head\_dim $<$ 32 on B200},
  year = {2026},
  url = {https://github.com/Dao-AILab/flash-attention/issues/2591},
}

@misc{flashattention-issue-2623,
  author = {FlashAttention},
  title = {[CuTe/SM90] Block-sparse backward returns silently-wrong gradients for head\_dim=64 (forward is exact)},
  year = {2026},
  url = {https://github.com/Dao-AILab/flash-attention/issues/2623},
}

@misc{flashattention-issue-1141,
  author = {FlashAttention},
  title = {The impact of setting head dim to an unconventional value on the algorithm's running speed},
  year = {2024},
  url = {https://github.com/Dao-AILab/flash-attention/issues/1141},
}

@misc{flashattention-issue-1476,
  author = {FlashAttention},
  title = {FA3 build error with large-dim changes},
  year = {2025},
  url = {https://github.com/Dao-AILab/flash-attention/issues/1476},
}

@misc{flashattention-issue-215,
  author = {FlashAttention},
  title = {The installation process always freezes after execution},
  year = {2023},
  url = {https://github.com/Dao-AILab/flash-attention/issues/215},
}

@misc{flashattention-issue-217,
  author = {FlashAttention},
  title = {IndexError about flash\_attn\_unpadded\_qkvpacked\_func},
  year = {2023},
  url = {https://github.com/Dao-AILab/flash-attention/issues/217},
}

@misc{flashattention-issue-385,
  author = {FlashAttention},
  title = {There seems to be errors in a small number of elements of dk in flash-attention-2},
  year = {2023},
  url = {https://github.com/Dao-AILab/flash-attention/issues/385},
}

@misc{flashattention-issue-282,
  author = {FlashAttention},
  title = {causal argument when q\_sequence is not cu\_seqlens\_q isn't aligned with cu\_seqlens\_k},
  year = {2023},
  url = {https://github.com/Dao-AILab/flash-attention/issues/282},
}

@misc{bitsandbytes-issue-444,
  author = {bitsandbytes},
  title = {ERROR: Your GPU does not support Int8 Matmul! NVIDIA V100},
  year = {2023},
  url = {https://github.com/bitsandbytes-foundation/bitsandbytes/issues/444},
}

@misc{bitsandbytes-issue-767,
  author = {bitsandbytes},
  title = {state.idx never gets reset in Linear8bitLt and MatMul8bitLt},
  year = {2023},
  url = {https://github.com/bitsandbytes-foundation/bitsandbytes/issues/767},
}

@misc{cutlass-issue-1801,
  author = {NVIDIA},
  title = {[BUG] TMA Cooperative GeMM with Stream-K scheduler hangs for specific gemm shapes},
  year = {2024},
  url = {https://github.com/NVIDIA/cutlass/issues/1801},
}

@misc{cutlass-issue-1270,
  author = {NVIDIA},
  title = {[BUG] Sparse GEMM seems to produce wrong results for F32 datatype and specific inputs},
  year = {2023},
  url = {https://github.com/NVIDIA/cutlass/issues/1270},
}

@misc{cutlass-issue-225,
  author = {NVIDIA},
  title = {15\_ampere\_sparse\_tensorop\_gemm fails},
  year = {2021},
  url = {https://github.com/NVIDIA/cutlass/issues/225},
}

@misc{cutlass-issue-1268,
  author = {NVIDIA},
  title = {[BUG] rs\_op\_selector does not allow to use mixed FP8 types for A and B},
  year = {2023},
  url = {https://github.com/NVIDIA/cutlass/issues/1268},
}

@misc{cutlass-issue-2494,
  author = {NVIDIA},
  title = {[BUG] Kernel Launch GridDim was miscalculated in examples/cute/tutorial/blackwell/04\_mma\_tma\_2sm\_sm100.cu},
  year = {2025},
  url = {https://github.com/NVIDIA/cutlass/issues/2494},
}

@misc{cutlass-issue-80,
  author = {NVIDIA},
  title = {08\_turing\_tensorop\_gemm test},
  year = {2020},
  url = {https://github.com/NVIDIA/cutlass/issues/80},
}

@misc{triton-issue-6480,
  author = {Triton},
  title = {Triton compile takes extremely long at runtime},
  year = {2025},
  url = {https://github.com/triton-lang/triton/issues/6480},
}

@misc{triton-issue-10596,
  author = {Triton},
  title = {cuda graph stream invalidate by sync call inside do\_bench\_cudagraph},
  year = {2026},
  url = {https://github.com/triton-lang/triton/issues/10596},
}

@misc{xformers-issue-719,
  author = {Meta},
  title = {[0.0.18] memory\_efficient\_attention NaNs when seqlen $>$ 32768},
  year = {2023},
  url = {https://github.com/facebookresearch/xformers/issues/719},
}

@misc{xformers-fix-719,
  author = {Meta},
  title = {[cutlassF] Fix K sequence length overflow},
  year = {2023},
  url = {https://github.com/facebookresearch/xformers/commit/68dce697013d3078315174840106bc9a81477d82},
}

@misc{deepgemm-issue-190,
  author = {DeepSeek},
  title = {[Bug]TMA Multicast code issue},
  year = {2025},
  url = {https://github.com/deepseek-ai/DeepGEMM/issues/190},
}

@misc{deepgemm-fix-193,
  author = {DeepSeek},
  title = {Fix multicast bug and optimize masked GEMM},
  year = {2025},
  url = {https://github.com/deepseek-ai/DeepGEMM/pull/193},
}

@misc{deepgemm-issue-147,
  author = {DeepSeek},
  title = {[bug] Unspecified Launch Failure of Normal GEMM when n=16},
  year = {2025},
  url = {https://github.com/deepseek-ai/DeepGEMM/issues/147},
}

@misc{deepgemm-fix-149,
  author = {DeepSeek},
  title = {Fix SM90 GEMM},
  year = {2025},
  url = {https://github.com/deepseek-ai/DeepGEMM/pull/149},
}

@inproceedings{Humbatova2020DLFaults,
  author = {Humbatova, Nargiz and Jahangirova, Gunel and Bavota, Gabriele and Riccio, Vincenzo and Stocco, Andrea and Tonella, Paolo},
  title = {Taxonomy of Real Faults in Deep Learning Systems},
  booktitle = {Proceedings of the ACM/IEEE 42nd International Conference on Software Engineering},
  year = {2020},
  pages = {1110--1121},
  publisher = {ACM},
  doi = {10.1145/3377811.3380395},
  url = {https://doi.org/10.1145/3377811.3380395}
}

@inproceedings{Betts2012GPUVerify,
  author = {Betts, Adam and Chong, Nathan and Donaldson, Alastair F. and Qadeer, Shaz and Thomson, Paul},
  title = {{GPUVerify}: A Verifier for {GPU} Kernels},
  booktitle = {Proceedings of the ACM International Conference on Object Oriented Programming Systems Languages and Applications},
  year = {2012},
  pages = {113--132},
  publisher = {ACM},
  doi = {10.1145/2384616.2384625},
  url = {https://doi.org/10.1145/2384616.2384625}
}

@inproceedings{Li2012GKLEE,
  author = {Li, Guodong and Li, Peng and Sawaya, Geof and Gopalakrishnan, Ganesh and Ghosh, Indradeep and Rajan, Sreeranga P.},
  title = {{GKLEE}: Concolic Verification and Test Generation for {GPU}s},
  booktitle = {Proceedings of the 17th ACM SIGPLAN Symposium on Principles and Practice of Parallel Programming},
  year = {2012},
  pages = {215--224},
  publisher = {ACM},
  doi = {10.1145/2145816.2145844},
  url = {https://doi.org/10.1145/2145816.2145844}
}

@inproceedings{Jiang2020CUDAsmith,
  author = {Jiang, Bo and Wang, Xiaoyan and Chan, Wing Kwong and Tse, T. H. and Li, Na and Yin, Yongfeng and Zhang, Zhenyu},
  title = {{CUDAsmith}: A Fuzzer for {CUDA} Compilers},
  booktitle = {2020 IEEE 44th Annual Computers, Software, and Applications Conference},
  year = {2020},
  pages = {861--871},
  publisher = {IEEE},
  doi = {10.1109/COMPSAC48688.2020.0-156},
  url = {https://doi.org/10.1109/COMPSAC48688.2020.0-156}
}

@inproceedings{Liu2023NNSmith,
  author = {Liu, Jiawei and Lin, Jinkun and Ruffy, Fabian and Tan, Cheng and Li, Jinyang and Panda, Aurojit and Zhang, Lingming},
  title = {{NNSmith}: Generating Diverse and Valid Test Cases for Deep Learning Compilers},
  booktitle = {Proceedings of the 28th ACM International Conference on Architectural Support for Programming Languages and Operating Systems, Volume 2},
  year = {2023},
  pages = {530--543},
  publisher = {ACM},
  doi = {10.1145/3575693.3575707},
  url = {https://doi.org/10.1145/3575693.3575707}
}

@inproceedings{Odena2019TensorFuzz,
  author = {Odena, Augustus and Olsson, Catherine and Andersen, David and Goodfellow, Ian},
  title = {{TensorFuzz}: Debugging Neural Networks with Coverage-Guided Fuzzing},
  booktitle = {Proceedings of the 36th International Conference on Machine Learning},
  series = {Proceedings of Machine Learning Research},
  volume = {97},
  year = {2019},
  pages = {4901--4911},
  publisher = {PMLR},
  url = {https://proceedings.mlr.press/v97/odena19a.html}
}

@inproceedings{Panchekha2015Herbie,
  author = {Panchekha, Pavel and Sanchez-Stern, Alex and Wilcox, James R. and Tatlock, Zachary},
  title = {Automatically Improving Accuracy for Floating Point Expressions},
  booktitle = {Proceedings of the 36th ACM SIGPLAN Conference on Programming Language Design and Implementation},
  year = {2015},
  pages = {1--11},
  publisher = {ACM},
  doi = {10.1145/2737924.2737959},
  url = {https://doi.org/10.1145/2737924.2737959}
}

@inproceedings{Tillet2019Triton,
  author = {Tillet, Philippe and Kung, H. T. and Cox, David},
  title = {{Triton}: An Intermediate Language and Compiler for Tiled Neural Network Computations},
  booktitle = {Proceedings of the 3rd ACM SIGPLAN International Workshop on Machine Learning and Programming Languages},
  year = {2019},
  pages = {10--19},
  publisher = {ACM},
  doi = {10.1145/3315508.3329973},
  url = {https://doi.org/10.1145/3315508.3329973}
}

@inproceedings{Peng2018CURD,
  author = {Peng, Yuanfeng and Grover, Vinod and Devietti, Joseph},
  title = {{CURD}: A Dynamic {CUDA} Race Detector},
  booktitle = {Proceedings of the 39th ACM SIGPLAN Conference on Programming Language Design and Implementation},
  year = {2018},
  pages = {390--403},
  publisher = {ACM},
  doi = {10.1145/3192366.3192368},
  url = {https://doi.org/10.1145/3192366.3192368}
}

@inproceedings{Chakraborty2025GPUMC,
  author = {Chakraborty, Soham and Krishna, S. and Pavlogiannis, Andreas and Tuppe, Omkar},
  title = {{GPUMC}: A Stateless Model Checker for {GPU} Weak Memory Concurrency},
  booktitle = {Computer Aided Verification},
  series = {Lecture Notes in Computer Science},
  year = {2025},
  pages = {321--346},
  publisher = {Springer Nature Switzerland},
  doi = {10.1007/978-3-031-98682-6_17},
  url = {https://doi.org/10.1007/978-3-031-98682-6_17}
}

@inproceedings{Zheng2020Ansor,
  author = {Zheng, Lianmin and Jia, Chengfan and Sun, Minmin and Wu, Zhao and Yu, Cody Hao and Haj-Ali, Ameer and Wang, Yida and Yang, Jun and Zhuo, Danyang and Sen, Koushik and Gonzalez, Joseph E. and Stoica, Ion},
  title = {Ansor: Generating {High-Performance} Tensor Programs for Deep Learning},
  booktitle = {14th USENIX Symposium on Operating Systems Design and Implementation},
  year = {2020},
  pages = {863--879},
  publisher = {USENIX Association},
  isbn = {978-1-939133-19-9},
  url = {https://www.usenix.org/conference/osdi20/presentation/zheng}
}

@inproceedings{Dao2022FlashAttention,
  author = {Dao, Tri and Fu, Dan and Ermon, Stefano and Rudra, Atri and R{\'e}, Christopher},
  title = {{FlashAttention}: Fast and Memory-Efficient Exact Attention with {IO}-Awareness},
  booktitle = {Advances in Neural Information Processing Systems},
  volume = {35},
  year = {2022},
  pages = {16344--16359},
  publisher = {Neural Information Processing Systems Foundation},
  doi = {10.52202/068431-1189},
  url = {https://doi.org/10.52202/068431-1189}
}

@misc{FlashInferRepo,
  author = {{FlashInfer contributors}},
  title = {{FlashInfer}},
  howpublished = {\url{https://github.com/flashinfer-ai/flashinfer}},
  year = {2026},
  note = {Accessed: 2026-08-10}
}

@misc{CUTLASSRepo,
  author = {{NVIDIA}},
  title = {{CUTLASS}},
  howpublished = {\url{https://github.com/NVIDIA/cutlass}},
  year = {2026},
  note = {Accessed: 2026-08-10}
}

@misc{DeepGEMMRepo,
  author = {{DeepSeek-AI}},
  title = {{DeepGEMM}},
  howpublished = {\url{https://github.com/deepseek-ai/DeepGEMM}},
  year = {2026},
  note = {Accessed: 2026-08-10}
}

@misc{XFormersRepo,
  author = {{Meta AI}},
  title = {{xFormers}},
  howpublished = {\url{https://github.com/facebookresearch/xformers}},
  year = {2026},
  note = {Accessed: 2026-08-10}
}

@misc{BitsAndBytesRepo,
  author = {{bitsandbytes contributors}},
  title = {{bitsandbytes}},
  howpublished = {\url{https://github.com/bitsandbytes-foundation/bitsandbytes}},
  year = {2026},
  note = {Accessed: 2026-08-10}
}

@misc{VLLMRepo,
  author = {{vLLM contributors}},
  title = {{vLLM}},
  howpublished = {\url{https://github.com/vllm-project/vllm}},
  year = {2026},
  note = {Accessed: 2026-08-10}
}

@misc{SGLangRepo,
  author = {{SGLang contributors}},
  title = {{SGLang}},
  howpublished = {\url{https://github.com/sgl-project/sglang}},
  year = {2026},
  note = {Accessed: 2026-08-10}
}

@inproceedings{Perry1985InterfaceFaults,
  author = {Perry, Dewayne E. and Evangelist, W. Michael},
  title = {An Empirical Study of Software Interface Faults},
  booktitle = {International Symposium on New Directions in Computing},
  year = {1985},
  pages = {32--38},
  publisher = {IEEE Computer Society},
  url = {https://users.ece.utexas.edu/~perry/work/papers/isnd.pdf}
}

@misc{nvidiacutedsl,
  author = {NVIDIA},
  title = {{CuTe DSL} Introduction},
  year = {2026},
  url = {https://docs.nvidia.com/cutlass/latest/media/docs/pythonDSL/cute_dsl_general/dsl_introduction.html},
  note = {Accessed: 2026-10-23}
}

\end{document}